\documentclass{article} 
\usepackage{arxiv_preprint,times}

\usepackage{amsmath,amsfonts,bm}

\def\eqref#1{equation~\ref{#1}}

\def\1{\bm{1}}

\DeclareMathAlphabet{\mathsfit}{\encodingdefault}{\sfdefault}{m}{sl}
\SetMathAlphabet{\mathsfit}{bold}{\encodingdefault}{\sfdefault}{bx}{n}

\usepackage{hyperref}
\usepackage{url}
\usepackage{graphicx}
\usepackage{booktabs}
\usepackage{multirow}
\usepackage{cleveref}
\usepackage[table]{xcolor}
\definecolor{subrotblue}{RGB}{235,243,250}
\newcommand{\subrotrow}[7]{%
  & & \cellcolor{subrotblue}#1
  & \cellcolor{subrotblue}#2
  & \cellcolor{subrotblue}#3
  & \cellcolor{subrotblue}#4
  & \cellcolor{subrotblue}#5
  & \cellcolor{subrotblue}#6
  & \cellcolor{subrotblue}#7 \\
}

\title{SubRot: Signed Gradient Subspace Calibration for VLM Rotation Quantization}

\author{
\textbf{Zhenhao Shang$^{1}$, Haizhao Jing$^{1}$, Haokui Zhang$^{1}$, Guoting Wei$^{2}$, Rong Xiao$^{3}$,}\\
\textbf{ Jianqing Gao$^{4}$, Peng Wang$^{1}$}\\[0.4em]
\textnormal{$^{1}$Northwest Polytechnical University}\\
\textnormal{$^{2}$Nanjing University of Science and Technology}\\
\textnormal{$^{3}$Intellifusion}\\
\textnormal{$^{4}$iFLYTEK CO., LTD}
}

\begin{document}

\maketitle

\begin{abstract}

Post-training quantization reduces the deployment cost of vision-language models (VLMs), but preserving multimodal capabilities at low bit widths remains challenging. Existing methods rely on modality- or token-level gradient statistics, which are susceptible to cross-sample variations in visual-to-textual token ratios and the positions of visual information, limiting statistical stability. Moreover, overly coarse aggregation through absolute values and averaging discards gradient signs and channel-wise differences, limiting the separation of modality-specific sensitivities. In contrast, the channel space provides a shared coordinate system across samples, making it a more natural basis for capturing stable task-sensitive structures. We therefore propose SubRot, a signed gradient subspace calibration method for VLM rotation quantization. Through eigendecomposition of the empirical Fisher matrix of activation gradients, SubRot identifies a sensitive channel subspace with three properties: cross-sample stability, clear sensitivity separation, and consistent signed effects on the autoregressive loss along certain directions. Guided by a local Taylor expansion, SubRot combines signed first-order guidance along sign-stable directions with second-order constraints along the remaining sensitive directions, while retaining MSE for overall reconstruction quality. This objective steers quantization errors toward loss-decreasing directions while controlling their magnitude. Experiments on five VLMs across five benchmarks show consistent average-score improvements over FlatQuant under W4A6 and W4A4, reaching 1.4 percentage points on LLaVA-NeXT-7B. Under W4A4, average accuracy degradation from FP16 remains within 1.4 percentage points across all evaluated models, while LLaVA-v1.5-13B exceeds its FP16 average score by 0.4 percentage points.
\end{abstract}

\section{Introduction}
Vision-language models (VLMs) have recently achieved remarkable progress in a wide range of tasks, including visual question answering, image captioning, document understanding, and multimodal reasoning, and are becoming an increasingly important foundation for general-purpose multimodal applications. However, state-of-the-art VLMs typically contain billions of parameters and process large numbers of visual tokens, resulting in substantial storage, memory-access, and computational costs during inference. These costs considerably hinder their deployment on resource-constrained devices and in latency-sensitive scenarios. Low-bit post-training quantization (PTQ), which reduces model storage and inference costs without requiring model retraining, has therefore become an important technique for efficient VLM deployment. 


Nevertheless, compared with large language models whose input distributions are relatively homogeneous, VLMs require the joint calibration of visual and textual tokens with substantially different activation magnitudes and distributions, making quantization calibration more challenging. Directly applying LLM quantization methods to VLMs overlooks this heterogeneous activation importance and can therefore lead to suboptimal calibration. Existing VLM quantization methods address this issue through modality-level reweighting\citep{li2025mbq}, modality-specific calibration\citep{yu2025mquant}, and token-level separation\citep{xiang2026fine}. However, their modality- or token-level gradient statistics face limitations in both stability and sensitivity separation. First, visual-to-textual token ratios and the positions of visual information vary substantially across samples, making these statistics dependent on sample-specific token compositions and limiting their cross-sample stability. Second, reducing gradients to scalar importance scores through absolute values and averaging introduces overly coarse aggregation which limits separability: taking absolute values discards the signed effects of activation errors on the autoregressive loss, while averaging across channels obscures differences in sensitivity. Moreover, these methods still aim solely to approximate the outputs of the original FP16 model, which theoretically constrains the performance ceiling attainable by the quantized model.

In this work, we employ the empirical Fisher matrix as a tractable approximation to the local curvature of the autoregressive loss with respect to intermediate activations, and perform spectral decomposition to identify a gradient-sensitive subspace along the channel dimension. Our empirical observations reveal three desirable properties of this subspace. First, the Fisher spectrum provides a clear separation between sensitive and insensitive subspaces, whose perturbations have substantially different effects on the autoregressive loss. Second, the extracted subspace structure and its sensitivity relationships remain stable across different calibration samples. Third, by further incorporating the signs of first-order gradients, we find that certain directions within the sensitive subspace exhibit consistent effects on the loss: within a local perturbation region, orienting quantization errors along specific directions can consistently lead to reductions in the autoregressive loss.

Building on these observations, we derive a new optimization objective for rotation quantization from the Taylor expansion of the autoregressive loss with respect to activation quantization errors, jointly modeling their signed first-order effects and second-order sensitivities. Specifically, a Fisher-based curvature term suppresses excessive quantization errors along sensitive directions, preventing rapid accumulation of second-order loss increases. Meanwhile, a signed first-order term encourages unavoidable quantization errors to align with directions that decrease the autoregressive loss. Consequently, our method goes beyond merely minimizing the overall magnitude of quantization errors and instead performs fine-grained error shaping within task-relevant subspaces, ensuring that the errors are magnitude-controlled and locally oriented in favorable directions. 

Experiments on five VLMs across five benchmarks show that SubRot consistently outperforms FlatQuant in average accuracy, with gains of up to 1.4 percentage points. Under W4A4, SubRot limits average accuracy degradation from FP16 to at most 1.4 percentage points across all evaluated models, while exceeding FP16 by 0.4 percentage points on LLaVA-v1.5-13B.

\section{Related Works}
\textbf{Post-training Quantization for LLMs.} 
A major line of research mitigates activation outliers through mathematically equivalent channel-wise transformations. SmoothQuant migrates the quantization difficulty from activations to weights through per-channel scaling, enabling hardware-efficient weight–activation quantization~\citep{xiao2023smoothquant}. AWQ instead identifies salient weights according to activation statistics and searches for channel-wise scaling factors that preserve these important weights under low-bit weight-only quantization~\citep{lin2024awq}. Although effective, such diagonal transformations have limited capability to redistribute complex outliers across channels.

Rotation-based methods extend this idea by mixing the channel dimensions before quantization. QuaRot applies computationally invariant Hadamard rotations to suppress hidden-state outliers and enables end-to-end 4-bit quantization of weights, activations, and KV caches~\citep{ashkboos2024quarot}. Moving beyond predefined rotations, SpinQuant optimizes orthogonal rotation matrices using calibration data to directly improve quantized-model accuracy~\citep{liu2025spinquant}. FlatQuant further replaces orthogonal rotations with learnable affine transformations for individual linear layers and adopts Kronecker decomposition to balance transformation capacity and runtime efficiency~\citep{sun2024flatquant}. 

Another family of methods exploits second-order or task-level information during quantization. GPTQ approximately minimizes the layer-wise output perturbation using an input-activation Hessian and sequential error compensation~\citep{frantar2022gptq}. GPTAQ introduces asymmetric calibration, matching each layer receiving quantized inputs against the corresponding output of the full-precision model to mitigate accumulated inter-layer errors~\citep{li2025gptaq}. GuidedQuant incorporates gradients of the end loss into the local quantization objective while preserving dependencies among weights within each output channel~\citep{kim2025guidedquant}. 

\textbf{Quantization for Vision-Language Models.}
Compared with text-only LLMs, vision-language models process heterogeneous visual and textual tokens whose numbers, activation distributions, and sensitivities differ substantially. VLMQ identifies visual-token redundancy and the distributional gap between visual and textual tokens, and introduces gradient-derived token importance factors into the Hessian-based weight-quantization objective~\citep{xue2025vlmq}. 

Subsequent methods focus more explicitly on heterogeneous activation distributions and quantization sensitivities. MBQ reveals that language and vision tokens exhibit substantially different loss sensitivities and reweights their reconstruction errors using modality-level gradient statistics during calibration~\citep{li2025mbq}. MQuant assigns separate static activation scales to visual and textual tokens and combines token reordering with rotation-magnitude suppression to support efficient fully static quantization~\citep{yu2025mquant}. Going beyond modality-level weighting, QIG employs quantization-aware integrated gradients to estimate fine-grained token sensitivity and uses the resulting importance scores to reweight channel-wise equalization objectives~\citep{xiang2026fine}. Most closely related to our channel-space perspective, C-PTQ estimates channel-wise sensitivity using a diagonal empirical Fisher approximation and penalizes each channel’s reconstruction residual according to its gradient variance during channel-wise scaling~\citep{li2026c}. Differing from C-PTQ, our method models correlated channel-sensitive subspaces through Fisher eigendecomposition and incorporates signed first-order information to guide rotation-based quantization.

\section{Method}
In this section, we first motivate our method by analyzing the limitations of existing gradient statistics methods and identifying potential improvements. We then introduce SubRot, detailing how to capture sensitive subspaces during calibration and leverage them to guide quantization calibration.
\subsection{Motivation}
Based on our analysis of existing methods, we identify two fundamental questions that gradient statistics must address to guide VLM quantization calibration: whether they can distinguish components with different magnitudes of impact on the task loss, and whether this distinction generalizes to samples not used for calibration. We therefore analyze existing methods in terms of separability and stability.

\textbf{Separability.}
Mainstream activation reweighting methods can be broadly formulated as follows. For activations of shape \(N \times C\), \(\boldsymbol{\epsilon}\) denotes the activation error between the full-precision and quantized models, \(a\) denotes the importance weight, \(\rho(\cdot)\) denotes the MAE or MSE metric, and \(\mathbf g\) denotes the corresponding gradient. When \(a_t\) is shared by all tokens within the same modality, the gradient statistics correspond to MBQ, when \(a_t\) varies across tokens, they correspond to QIG.
\begin{equation}
    \mathcal {J} _ {\mathrm {scalar}} = \sum_ {t = 1} ^ {N} a _ {t} \rho \left(\boldsymbol {\epsilon} _ {t}\right), \quad 
    a _ {t} = \frac {1}{C} \sum_ {c = 1} ^ {C} \left| g _ {t, c} \right|,
\end{equation}
However, when computing the importance weight \(a\), taking the absolute value ignores how the gradient sign affects the direction of change in the model’s autoregressive loss, while averaging across channels obscures channel-wise differences in importance. Consider the following example:
\begin{equation}
    \mathbf{if}\: g_1 = (3, -1)^T,\quad g_2 = (2, 2)^T, \quad a_1 = a_2 = 2
\end{equation}
\begin{equation}
    \mathbf{set}\: \epsilon = (\delta, 3\delta),\quad \Delta \mathcal{L}_t\approx \mathbf{g}_t^{\top}\epsilon_t,\quad \Delta L_1 \approx 0,\quad \Delta L_2 \approx 8\delta
\end{equation}
As shown, the importance scores obtained by existing gradient-based statistics may fail, in certain cases, to accurately measure the effect of activation errors on the model’s autoregressive loss, resulting in a theoretical limitation in separability. Therefore, in this work, we explore an approach that preserves the original gradient information to the greatest extent possible, without relying on heuristic high-dimensional aggregation.

\textbf{Stability.} VLM quantization typically relies on only a small calibration set to estimate importance and search for quantization parameters. Therefore, an effective gradient statistic should not only accurately characterize sensitivity on the calibration samples but also remain consistent across samples, enabling generalization to unseen inputs. However, existing methods mainly aggregate gradient information along the modality or token dimension. Such token-indexed statistics lack stable cross-sample correspondence, as different samples typically vary in the number of visual tokens, text length, and prompt structure. 
\begin{equation}
\text{the } t\text{-th token in } \mathbf{x}^{(s)}
\not\equiv
\text{the } t\text{-th token in } \mathbf{x}^{(s')}.
\end{equation}
A stable importance measure should therefore be defined in a representation space with fixed dimensionality and explicit cross-sample correspondence, rather than depending heavily on sample-specific token compositions. In contrast, the channel space provides a unified coordinate system shared across samples, making it a more natural basis for accumulating stable task-sensitive structures.

\begin{figure}[t]
    \centering
    \includegraphics[width=\linewidth]{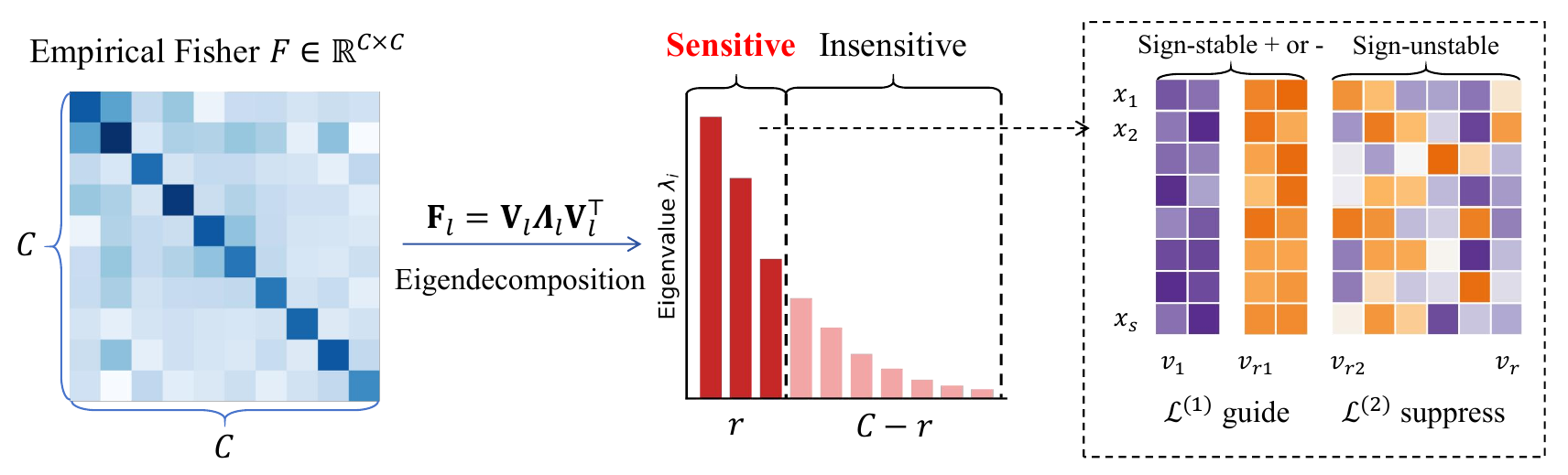}
    \caption{\textbf{Overview of the proposed Fisher-informed subspace calibration. }We eigendecompose the empirical Fisher matrix and select the top-\(r\) eigen-directions as the sensitive subspace. Within this subspace, sign-stable directions are guided by the signed first-order loss \(\mathcal{L}^{(1)}\), while sign-unstable directions are constrained by the second-order loss \(\mathcal{L}^{(2)}\) to suppress quantization errors.}
    \label{fig:method}
\end{figure}

\subsection{Capturing sensitive subspaces}
First, for the \(l\)-th layer to be quantized, let \(\mathbf{Y}^{(l)}\) and \(\widehat{\mathbf{Y}}^{(l)}\) denote its full-precision and quantized output activations, respectively. We perform a Taylor expansion of the model’s autoregressive loss around \(\mathbf{Y}^{(l)}\), while omitting the higher-order terms:
\begin{equation}
\Delta \mathcal{L}_l = \mathcal{L}(\widehat{\mathbf{Y}}_l) - \mathcal{L}(\mathbf{Y}_l) 
\approx \langle \mathbf{G}_l, \Delta \mathbf{Y}_l \rangle_F + \frac{1}{2} \operatorname{vec}(\Delta \mathbf{Y}_l)^\top \mathbf{H}_l \operatorname{vec}(\Delta \mathbf{Y}_l),
\end{equation}

Here, \(\mathbf{G}=\nabla_{\mathbf{Y}^{(l)}}\mathcal{L}\) and \(\mathbf{H}=\nabla_{\mathbf{Y}^{(l)}}^{2}\mathcal{L}\). Since explicitly constructing the full Hessian matrix is computationally prohibitive, we employ the empirical Fisher matrix as a tractable positive-semidefinite proxy for the local curvature. Given a calibration set of \(S\) samples, the channel-wise empirical Fisher matrix accumulated over all valid tokens can be expressed as follows:
\begin{equation}
    \mathbf{H}_l \approx \mathbf {F} _ {l} = \frac {1}{Z _ {l}} \sum_ {s = 1} ^ {S} \left(\mathbf {G} _ {l} ^ {(s)}\right) ^ {\top} \mathbf {G} _ {l} ^ {(s)}, \quad Z _ {l} = \sum_ {s = 1} ^ {S} N _ {s}.
\end{equation}
Since \(\mathbf{F}\) is a symmetric positive-semidefinite matrix, we perform eigendecomposition on it. As illustrated in the figure \ref{fig:method}, the channel-wise empirical Fisher exhibits a highly concentrated eigenspectrum: a small number of principal eigen-directions account for most of the spectral energy, while the eigenvalues associated with the remaining directions decay rapidly. We therefore select the eigenvectors corresponding to the \(r\) largest eigenvalues, where \(r \ll C\), to construct the sensitive subspace basis \(\mathbf{U}_l\) for the \(l\)-th layer:
\begin{equation}
    \mathbf{F}_l = \mathbf{V}_l\boldsymbol{\Lambda}_l\mathbf{V}_l^\top,\quad \mathbf {U} _ {l} = \left[ \mathbf {v} _ {l, 1}, \mathbf {v} _ {l, 2}, \dots , \mathbf {v} _ {l, r} \right] \in \mathbb {R} ^ {C \times r}
\end{equation}


We find that the extracted sensitive subspaces exhibit three key properties: \textbf{cross-sample stability, clear separation between sensitive and insensitive directions, and stable signed first-order effects.} First, shared channel coordinates enable consistent cross-sample gradient aggregation. Second, a concentrated Fisher spectrum separates leading directions with high quadratic sensitivity from the remaining directions with low sensitivity. 
The first two properties are then analyzed in the experimental section.

For the third property, second-order weight-quantization methods represented by OBQ\citep{frantar2022OBQ} and GPTQ\citep{frantar2022gptq} typically rely on a local stationarity assumption in the parameter space: once training has converged, the gradients with respect to the weights are assumed to be approximately zero, and the second-order effects of quantization perturbations therefore receive primary attention. However, activations are not independently optimized parameters during training. Consequently, even when the model weights have converged, the first-order derivative of the loss with respect to an intermediate activation generally does not vanish. 
Since activation gradients are highly input-dependent, extracting consistent first-order information across different samples is challenging. 
Nevertheless, we find that the subspace contains cross-sample stable directions with consistent signed effects:
\begin{equation}
    \exists \mathbf{u}_k\in \mathcal{S}_{\mathrm{sens}}, \sigma_k\in \{-1, +1\}, \quad \Pr_i\left[\sigma_k\cdot \frac{1}{N_i}\sum_{n=1}^{N_i}\left(\nabla_{\mathbf{x}_{i,n}}\mathcal{L}_i\right)^\top\mathbf{u}_k>0\right]\approx 1.
\end{equation}
Where \(\mathbf{u}_k\) denotes a direction in the sensitive subspace, \(i\) indexes different samples, \(n\) indexes tokens, and \(\sigma_k\) denotes the stable sign of the first-order effect along this direction. Since the directional derivative along \(\mathbf{u}_k\) exhibits a consistent sign across the vast majority of samples, perturbations along this direction have a cross-sample stable increasing or decreasing effect on the autoregressive loss, enabling the first-order term to be reliably exploited during quantization calibration.

\subsection{Guiding Quantization Calibration Using Sensitive Subspaces}
We redesign the optimization objective used during quantization calibration. The widely adopted MSE reconstruction loss assigns equal importance to all activation errors, overlooking the fact that different directions in the channel space can have substantially different effects on the model’s autoregressive loss. Moreover, it merely encourages the quantized representations to approximate their FP16 counterparts as closely as possible. To address these limitations, we first project the discrepancy between the quantized and FP16 activations onto the task-sensitive subspace using the corresponding subspace basis. Based on a local Taylor expansion of the autoregressive loss with respect to the intermediate activations, we then introduce a signed first-order loss for directions whose first-order effects exhibit consistent signs across samples, thereby guiding the quantization error toward directions that reduce the autoregressive loss. Meanwhile, for the remaining sensitive directions with unstable first-order effects, we impose a second-order curvature constraint to suppress excessive quantization errors along highly sensitive directions. The resulting objective is formulated as follows.
\begin{equation}
    \mathcal{L}=\mathcal{L}_{\mathrm{MSE}}+\alpha\underbrace{\mathrm{Mean}\left(\Delta\mathbf{Z}_{\mathrm{st}}\odot\mathbf{\boldsymbol{\sigma}}\right)}_{\mathcal{L}_{\mathrm{first}}}+\beta\underbrace{\left\|\Delta\mathbf{Z}_{\mathrm{us}}\right\|_{F}^{2}}_{\mathcal{L}_{\mathrm{second}}}
\end{equation}
Here, \(\alpha\) and \(\beta\) are weighting coefficients; \(\Delta\mathbf{Z}_{\mathrm{st}}\) denotes the quantization errors along directions in the sensitive subspace with stable first-order effects, while \(\Delta\mathbf{Z}_{\mathrm{us}}\) denotes the errors along the remaining sensitive directions whose first-order signs are unstable across samples. \(\boldsymbol{\sigma}\) represents the sign term. The former determines the favorable direction of the quantization error, the latter controls its magnitude, and the original MSE loss preserves the overall reconstruction quality.

\begin{table}[t]
    \centering
    \caption{Quantization results of LLaVA-NeXT-7B, Qwen2-VL-7B, and LLaVA-v1.5-7B on five benchmarks under W4A6 and W4A4 settings, with FP16 performance as a reference.}
    \resizebox{\textwidth}{!}{
    \begin{tabular}{ccc ccccc c}
\toprule
Model & Bitwidth &  Method & MMMU & OCRBench & VizWiz & ChartQA & SEED-2+ & Average  \\
\midrule
\multirow{11}{*}{LLaVA-Next-7b}
& fp16 & - 
&36.7 	&51.2 	&58.7 	&50.4 	&50.6 	&49.5 \\
\cmidrule{2-9}
& \multirow{5}{*}{W4A6}
& RTN
&14.8 	&23.7 	&37.5 	&27.6 	&3.9 	&21.5 \\
&& MBQ
&28.8 	&40.6 	&52.9 	&41.7 	&41.1 	&41.0 \\
&& Quarot
&32.5 	&47.6 	&53.7 	&50.8 	&48.0 	&46.5 \\
&& FlatQuant 
&33.1 	&50.5 	&57.2 	&52.5 	&51.3 	&48.9 \\
\subrotrow{SubRot}{\textbf{35.0}}{\textbf{51.8}}{\textbf{58.1}}{\textbf{52.7}}{\textbf{51.4}}{\textbf{49.8}}    
\cmidrule{2-9}
& \multirow{5}{*}{W4A4}
& RTN
&4.6 	&0.4 	&0.1 	&0.2 	&4.8 	&2.0 \\
&& MBQ
&5.5 	&7.0 	&23.2 	&11.7 	&3.4 	&10.2 \\
&& Quarot
&29.8 	&45.9 	&53.2 	&48.0 	&45.1 	&44.4 \\
&& FlatQuant  
&\textbf{34.5} 	&48.9 	&54.7 	&49.0 	&49.8 	&47.4 \\
\subrotrow{SubRot}{33.3}{\textbf{50.3}}{\textbf{57.2}}{\textbf{52.4}}{\textbf{50.7}}{\textbf{48.8}}
\midrule

\multirow{11}{*}{Qwen2VL-7b}
& fp16 & -  
&51.5 	&77.6 	&68.2 	&80.9 	&69.5 	&69.5 \\
\cmidrule{2-9}
& \multirow{5}{*}{W4A6}
& RTN
&43.4 	&75.1 	&61.9 	&68.1 	&63.2 	&62.3 \\
&& MBQ
&41.7 	&62.9 	&45.8 	&72.3 	&65.3 	&57.6 \\
&& Quarot
&45.7 	&\textbf{77.2} 	&67.0 	&77.4 	&65.8 	&66.6 \\
&& FlatQuant  
&49.8 	&76.5 	&68.2 	&79.5 	&68.6 	&68.5 \\
\subrotrow{SubRot}{\textbf{53.7}}{76.4}{\textbf{68.4}}{\textbf{80.4}}{\textbf{68.8}}{\textbf{69.5}}
\cmidrule{2-9}
& \multirow{5}{*}{W4A4}
& RTN
&0.4 	&0.0 	&0.0 	&0.0 	&0.5 	&0.2 \\
&& MBQ
&5.3 	&0.3 	&0.0 	&0.1 	&3.5 	&1.8 \\
&& Quarot
&9.2 	&31.9 	&11.3 	&7.8 	&18.5 	&15.7 \\
&& FlatQuant  
&\textbf{50.5} 	&76.1 	&67.5 	&\textbf{79.4} 	&67.9 	&68.3 \\
\subrotrow{SubRot}{50.1}{\textbf{76.9}}{\textbf{68.4}}{79.3}{\textbf{68.5}}{\textbf{68.6}}
\midrule

\multirow{11}{*}{LLaVA-1.5-7b}
& fp16 & - 
&34.7 	&17.1 	&54.6 	&17.8 	&41.2 	&33.1 \\
\cmidrule{2-9}
& \multirow{5}{*}{W4A6}
& RTN
&4.4 	&3.6 	&2.5 	&1.4 	&1.8 	&2.7 \\
&& MBQ
&31.7 	&15.1 	&53.6 	&14.6 	&33.9 	&29.8 \\
&& Quarot
&31.5 	&16.9 	&55.2 	&17.1 	&38.5 	&31.8 \\
&& FlatQuant
&\textbf{34.7} 	&\textbf{17.1} 	&54.6 	&\textbf{17.4} 	&40.1 	&32.8 \\
\subrotrow{SubRot}{\textbf{34.7}}{\textbf{17.1}}{\textbf{55.8}}{17.1}{\textbf{40.4}}{\textbf{33.0}}
\cmidrule{2-9}
& \multirow{5}{*}{W4A4}
& RTN
&3.8 	&0.2 	&1.1 	&0.5 	&1.6 	&1.4 \\
&& MBQ
&4.5 	&1.5 	&1.6 	&0.2 	&2.6 	&2.1 \\
&& Quarot
&31.5 	&16.8 	&54.2 	&15.5 	&36.5 	&30.9 \\
&& FlatQuant
&\textbf{35.1} 	&16.8 	&53.7 	&17.1 	&40.1 	&32.6 \\
\subrotrow{SubRot}{34.7}{\textbf{17.1}}{\textbf{55.4}}{\textbf{17.3}}{\textbf{40.4}}{\textbf{33.0}}

\bottomrule

    \end{tabular}
    }
    \label{tab:main table 1}
\end{table}

\begin{table}[t]
    \centering
    \caption{Quantization results of Qwen2-VL-2B and LLaVA-v1.5-13B on five benchmarks under W4A6 and W4A4 settings, with FP16 performance as a reference.}
    \resizebox{\textwidth}{!}{
    \begin{tabular}{ccc ccccc c}
\toprule
Model & Bitwidth &  Method & MMMU & OCRBench & VizWiz & ChartQA & SEED-2+ & Average  \\
\midrule
\multirow{11}{*}{Qwen2VL-2b}
& fp16 & - 
&42.1 	&75.3 	&66.3 	&72.3 	&62.1 	&63.6 \\
\cmidrule{2-9}
& \multirow{5}{*}{W4A6}
& RTN
&36.8 	&68.2 	&59.4 	&60.0 	&56.2 	&56.1 \\
&& MBQ
&33.2 	&63.6 	&57.6 	&59.2 	&54.5 	&53.6 \\
&& Quarot
&36.7 	&71.4 	&57.1 	&66.5 	&59.3 	&58.2 \\
&& FlatQuant 
&39.3 	&75.3 	&65.0 	&71.5 	&\textbf{61.9} 	&62.6 \\
\subrotrow{SubRot}{\textbf{42.5}}{\textbf{75.8}}{\textbf{65.3}}{\textbf{71.8}}{61.4}{\textbf{63.4}}
\cmidrule{2-9}
& \multirow{5}{*}{W4A4}
& RTN
&4.4 	&6.8 	&0.1 	&0.0 	&4.2 	&3.1 \\
&& MBQ
&2.3 	&1.0 	&0.1 	&0.0 	&1.5 	&1.0 \\
&& Quarot
&11.9 	&35.9 	&11.8 	&15.8 	&18.0 	&18.7 \\
&& FlatQuant 
&40.6 	&\textbf{74.8} 	&64.5 	&69.6 	&61.0 	&62.1 \\
\subrotrow{SubRot}{\textbf{41.2}}{74.0}{\textbf{64.6}}{\textbf{69.7}}{\textbf{61.4}}{\textbf{62.2}}
\midrule
\multirow{11}{*}{LLaVA-1.5-13b}
& fp16 & - 
&36.5 	&20.0 	&59.1 	&19.1 	&44.0 	&35.7 \\
\cmidrule{2-9}
& \multirow{5}{*}{W4A6}
& RTN
&19.7 	&7.3 	&53.1 	&12.4 	&28.9 	&24.3 \\
&& MBQ
&35.0 	&18.6 	&57.0 	&15.1 	&40.0 	&33.1 \\
&& Quarot
&36.4 	&19.4 	&54.3 	&\textbf{19.4} 	&42.7 	&34.4 \\
&& FlatQuant 
&37.1 	&20.6 	&\textbf{60.0 }	&18.2 	&\textbf{44.4} 	&36.1 \\
\subrotrow{SubRot}{\textbf{38.1}}{\textbf{20.8}}{59.9}{18.3}{44.0}{\textbf{36.2}}
\cmidrule{2-9}
& \multirow{5}{*}{W4A4}
& RTN
&1.0 	&0.0 	&0.0 	&0.0 	&0.5 	&0.3 \\
&& MBQ
&3.5 	&0.0 	&0.0 	&0.0 	&2.8 	&1.3 \\
&& Quarot
&32.8 	&19.6 	&54.2 	&17.9 	&42.0 	&33.3 \\
&& FlatQuant 
&37.3 	&20.2 	&59.8 	&17.7 	&43.4 	&35.7 \\
\subrotrow{SubRot}{\textbf{38.1}}{\textbf{20.5}}{\textbf{59.9}}{\textbf{18.2}}{\textbf{43.6}}{\textbf{36.1}}
\bottomrule
    \end{tabular}
    }
    \label{tab:main table 2}
\end{table}

\begin{table}[t]
    \centering
    \caption{Ablation study of the first-order guidance loss for sign-stable directions and the second-order suppression loss for sensitive directions. Experiments are conducted on LLaVA-NeXT-7B under the W4A6 quantization setting.}
    \begin{tabular}{cccccc}
\toprule
$\mathcal{L}^{(1)}$ guide & $\mathcal{L}^{(2)}$ suppress& MMMU & OCRBench & SEED-2+ & Average($\uparrow$) \\
\midrule
- & -                   &33.1 	&50.5 	&51.3 &45.0\\
\checkmark &-           &34.4 	&50.8 	&50.6 &45.3 \\
-& \checkmark           &34.8 	&50.1 	&51.5 &45.5\\
\checkmark & \checkmark &35.0 	&51.8 	&51.4 &46.1\\

\bottomrule
    \end{tabular}
    \label{tab:ablation}
\end{table}


\begin{figure}[t]
    \centering
    \includegraphics[width=\linewidth]{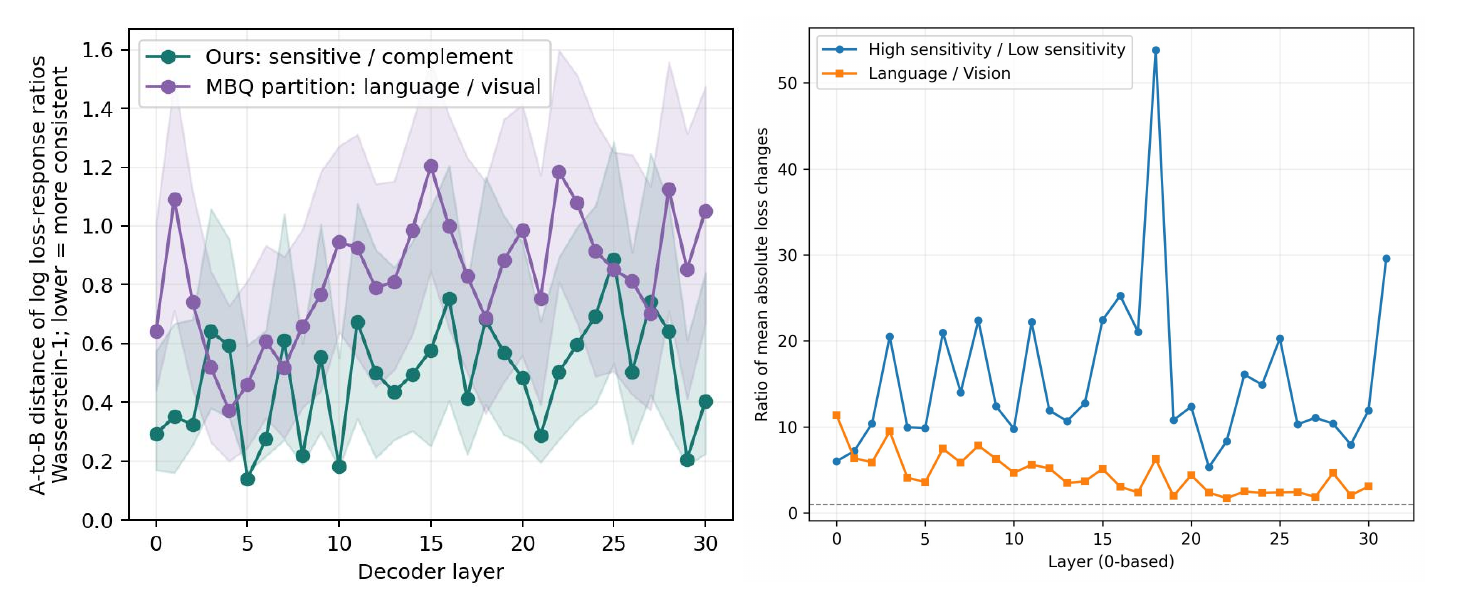}
    \caption{Left: Stability of importance partitions under distribution shift. The \(y\)-axis reports the Wasserstein-1 distance between the two log-ratio distributions, where lower values indicate greater stability. Shaded regions denote 95\% bootstrap confidence intervals. Right: Layer-wise ratios of mean absolute autoregressive loss changes.}
    \label{fig:feature1&2}
\end{figure}

\begin{figure}[t]
    \centering
    \includegraphics[width=\linewidth]{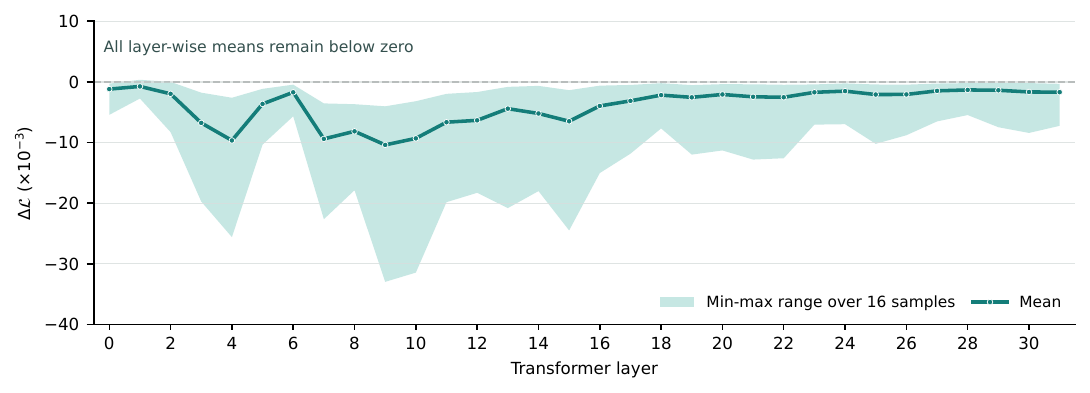}
    \caption{\textbf{Cross-sample stability of the signed first-order effect.} The solid line represents the sample mean, and the shaded region indicates the minimum–maximum range; negative values indicate a decrease in loss.}
    \label{fig:L1 stable}
\end{figure}

\section{Experiments}
\subsection{Experimental Setup}
\textbf{Data and Models.} We randomly select 128 samples from ScienceQA\citep{scienceQA} for calibration, providing multimodal task signals while keeping the calibration cost manageable. We evaluate accuracy on MMMU\citep{yue2024mmmu}, OCRBench\citep{liu2024ocrbench}, VizWiz\citep{gurari2018vizwiz}, ChartQA\citep{masry2022chartqa}, and SEEDBench2Plus\citep{li2024seed}, covering multidisciplinary understanding, text recognition, real-world visual question answering, chart comprehension, and fine-grained multimodal understanding. We conduct experiments on LLaVA-NeXT-7B\citep{liu2024llavanext}, Qwen2VL-7b\citep{wang2024qwen2}, LLaVA-v1.5-7b\citep{llava}, LLaVA-v1.5-13B, and Qwen2-VL-2B, spanning two model families and parameter scales from 2B to 13B to examine the applicability of our method across models and sizes.

\textbf{Baselines and Quantization Settings.} We compare our method with RTN, MBQ, QuaRot, and FlatQuant. RTN serves as a basic round-to-nearest quantization baseline, while MBQ represent multimodal-aware reweighting calibration. QuaRot and FlatQuant enable comparisons with rotation-based and learnable-transformation-based quantization methods. Quantization is applied to the language model component. We report the accuracy of the original FP16 models as a full-precision reference and evaluate W4A6 and W4A4 configurations, where W and A denote the bit widths of weights and activations, respectively. 

\subsection{Main Results}


Table \ref{tab:main table 1} and Table \ref{tab:main table 2} report the evaluation results of five vision-language models under the W4A6 and W4A4 settings. SubRot achieves the highest average score across various models and bit-width configurations, demonstrating its applicability across different model families and parameter scales. Compared with FlatQuant, SubRot yields cross-model average improvements of 0.60 and 0.52 percentage points under W4A6 and W4A4, respectively. In particular, on LLaVA-NeXT-7B, it improves the average score by 0.9 and 1.4 percentage points under the two respective settings.

Under the more challenging W4A4 setting, SubRot still effectively preserves model accuracy, while RTN and MBQ suffer severe performance degradation on multiple models. The average scores of QuaRot on Qwen2-VL-7B and Qwen2-VL-2B drop to 15.7 and 18.7, respectively, whereas SubRot achieves substantially higher scores of 68.6 and 62.2.

Compared with the FP16 models, SubRot limits the average-score gap to no more than 0.5 percentage points for all models under W4A6, while the corresponding degradation under W4A4 remains within 1.4 percentage points. Moreover, on LLaVA-v1.5-7B and LLaVA-v1.5-13B, SubRot further improves the average accuracy even when FlatQuant has already matched or surpassed the FP16 baseline. These results demonstrate that SubRot more effectively preserves multimodal task performance under low-bit quantization and generalizes well across different model families and parameter scales.

\subsection{Ablation Studies}
\subsubsection{Accuracy contribution of sensitive-subspace calibration.}
To evaluate the effectiveness of individual loss terms in sensitive-subspace calibration, we performed an ablation study on LLaVA-NeXT-7B under the W4A6 quantization setting in table \ref{tab:ablation}. Specifically, we separately examine the first-order guidance term for sign-stable directions and the second-order suppression term for the remaining sensitive directions. Performance is evaluated on MMMU, OCRBench, and SEEDBench2Plus.

Without either loss term, the model achieves an average score of 45.0. Introducing only the first-order guidance term or the second-order suppression term improves the average score to 45.3 and 45.5, respectively, demonstrating the benefits of both guiding the direction of quantization errors and constraining their magnitude along sensitive directions. Combining the two terms yields the highest average score of 46.1, outperforming the baseline by 1.1 percentage points. The results demonstrate that the two loss terms are complementary and jointly improve the accuracy of the quantized model.

\subsubsection{Cross-sample stability of the subspace}
To evaluate the stability of the importance partition identified during calibration under distribution shift, we conduct a controlled noise intervention experiment on LLaVA-NeXT-7B. We construct a calibration set \(A\) using 128 samples from ScienceQA, on which we fit the sensitive subspace and estimate the modality-level gradient weights used by MBQ. We then randomly select 64 samples from \(A\) to form the evaluation set \(A_{\mathrm{eval}}\), and construct another evaluation set \(B\) using 64 samples outside the calibration set with relatively high ratios of visual tokens to input-text tokens. For each decoder layer under evaluation, Gaussian noise is separately injected into the sensitive and non-sensitive subspaces, as well as the language and visual tokens, for both evaluation sets. We then examine the consistency of the resulting changes in the model’s autoregressive loss.

As shown in \cref{fig:feature1&2} left, the channel-subspace partition yields smaller cross-set distances across most decoder layers, indicating that the relative loss responses of its sensitive and non-sensitive partitions are more robust to the constructed input distribution shift. In contrast, the language/visual token partition exhibits substantially larger distributional changes in several middle and deeper layers.

\subsubsection{Separation between sensitive and insensitive directions}
To evaluate the separability of the sensitive subspace, we conduct a comparative noise perturbation experiment using two partitioning schemes: sensitive versus non-sensitive directions, and language versus visual tokens. For each scheme, noise of equal magnitude is separately applied to its two partitions, and the ratio between the resulting changes in the model’s autoregressive loss is measured. As shown in \cref{fig:feature1&2} right, the sensitive/non-sensitive partition exhibits greater separability than the language/visual token partition in the vast majority of layers, demonstrating the pronounced separation induced by the sensitive subspace.

\subsubsection{Stability of the signed first-order effect}
To verify the stability of the signed first-order effect in the sensitive subspace, we first identify the sign-stable directions and then apply fixed-magnitude perturbations according to their corresponding signs: negative perturbations are applied to consistently positive directions, while positive perturbations are applied to consistently negative directions. We subsequently perform inference on 16 randomly selected samples independent of the calibration set and measure the change in autoregressive loss before and after perturbation. As shown in \cref{fig:L1 stable}, despite variations across layers and samples, the loss changes remain negative for every sample across all layers, with no sign reversal that increases the loss. These results demonstrate that the signed first-order effects of certain directions in the sensitive subspace are stable and can therefore provide reliable guidance for quantization calibration.

\begin{table}[t]
    \centering
    \caption{End-to-end inference speed and peak GPU memory usage measured after actual quantized deployment of LLaVA-Next-7B on 4,319 samples from the VizWiz dataset with single RTX 4090.}
    \resizebox{0.7\linewidth}{!}{
    \begin{tabular}{cccc}
\toprule
Bitwidth & Method & End-to-end average time(ms) & Peak Memory(GB)  \\
\midrule
fp16    
&-          &2146.7 &14.9   \\
\midrule
\multirow{3}{*}{W4A4}    
&Quarot     &1588.9 &8.5    \\
&FlatQuant  &1303.8 &9.7    \\
&SubRot     &1303.8 &9.7    \\
\bottomrule
    \end{tabular}
    }
    \label{tab:infer time}
\end{table}

\subsection{Evaluation with Actual Quantized Deployment}
We evaluated inference speed and peak GPU memory usage under actual quantized deployment to assess the efficiency gains from quantization. Using LLaVA-Next-7B, We measured end-to-end inference time and peak GPU memory usage on 4,319 samples from VizWiz using a single NVIDIA RTX 4090 GPU and a batch size of 1. The table \ref{tab:infer time} compares the FP16 model with QuaRot, FlatQuant, and SubRot under W4A4 quantization. Compared with FP16, SubRot reduces the average end-to-end inference time from 2146.7 ms to 1303.8 ms, a 39.3\% reduction, and peak GPU memory usage from 14.9 GB to 9.7 GB, a 34.9\% reduction. Because SubRot does not alter FlatQuant’s inference computation graph, the two methods are expected to have comparable inference speed and memory usage, as confirmed by the results.

\section{Conclusion}
In this work, we revisit low-bit rotation quantization for VLMs from the channel-space perspective. By eigendecomposing the empirical Fisher matrix, we identify a task-sensitive subspace with cross-sample stability, clear sensitivity separation, and consistent signed effects. Based on these properties, SubRot combines second-order curvature suppression with first-order signed guidance to control quantization-error magnitudes and guide perturbations toward loss-decreasing directions. Extensive experiments demonstrate its effectiveness across different VLM families and model scales.

\subsection*{AI use statement}
In this work, we used generative AI tools to assist with language editing, including improving grammar, clarity, and presentation. We did not use generative AI tools to generate research ideas, develop the methodology, analyze results, write code, create figures, or identify citations. All AI-assisted text was carefully reviewed by the authors. We take full responsibility for the final content of this work, including any text produced with the aid of generative AI.

\bibliography{references}
\bibliographystyle{arxiv_preprint}

\appendix
\section{Appendix}
\subsection{Impact on calibration time}
To assess the impact of subspace statistics on calibration time, We randomly sampled 128 examples from ScienceQA as the calibration set for Qwen2-VL-7B and measured the time required to complete calibration with FlatQuant and SubRot using four NVIDIA RTX 4090 GPUs. For SubRot, we further divided the time into subspace statistics and search/training. As shown in \cref{tab:time cost}, the additional time is primarily attributable to subspace statistics, while the search/training time remains nearly unchanged.
\begin{table}[h]
    \centering
    \caption{Calibration time breakdown of FlatQuant and SubRot (mm:ss).}
    \resizebox{0.55\linewidth}{!}{
    \begin{tabular}{cccc}
\toprule
Method & Subspace Statistics & Search/Training & Total\\
\midrule
FlatQuant   & -     & 39:43     & 39:43\\
SubRot      &6:59   & 39:47     & 46:46\\
\bottomrule
    \end{tabular}
    }
    
    \label{tab:time cost}
\end{table}

\begin{figure}[t]
    \centering
    \setlength{\tabcolsep}{1pt} 
    \renewcommand{\arraystretch}{0} 
    \begin{tabular}{@{}cccc@{}}
        \includegraphics[width=.235\columnwidth]{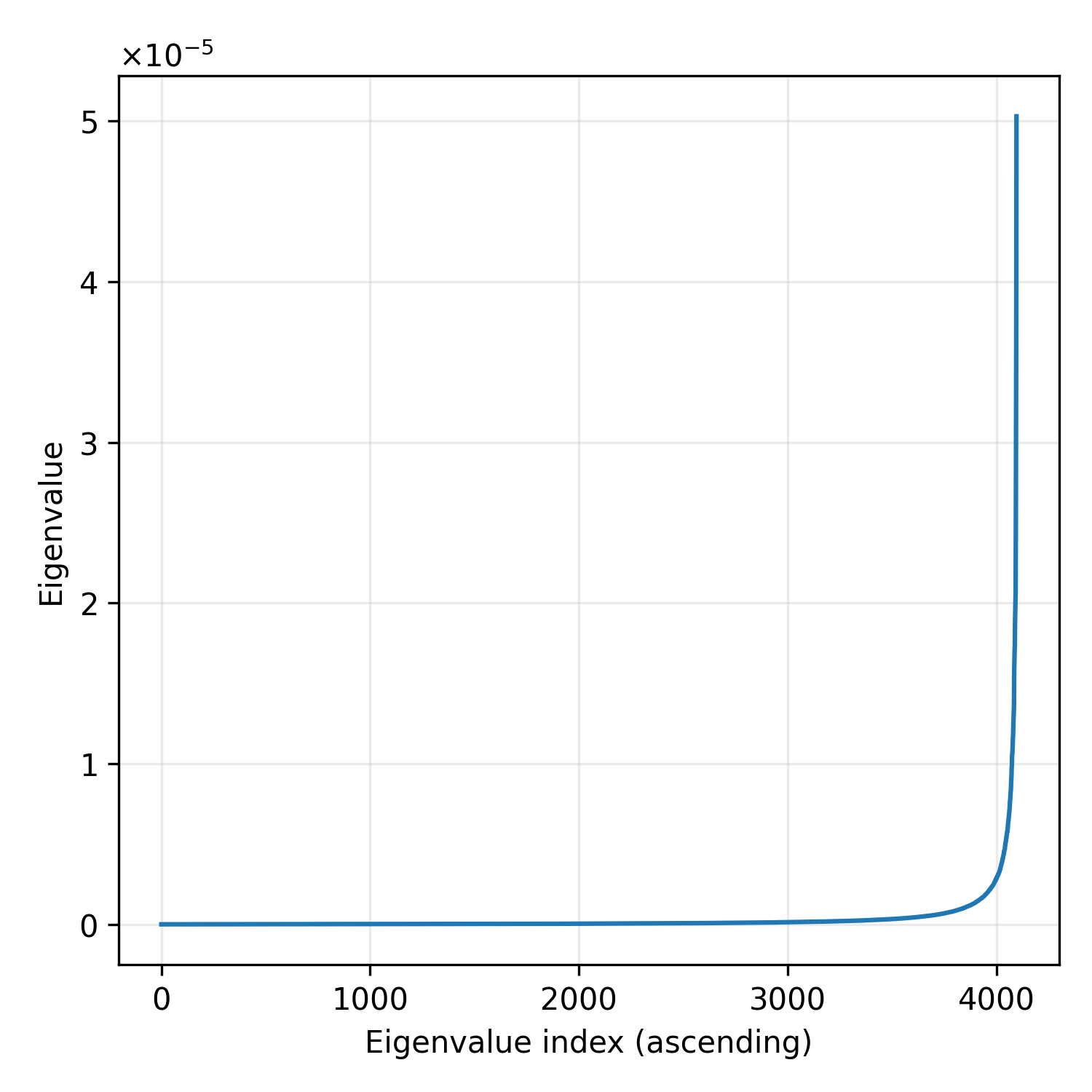} &
        \includegraphics[width=.235\columnwidth]{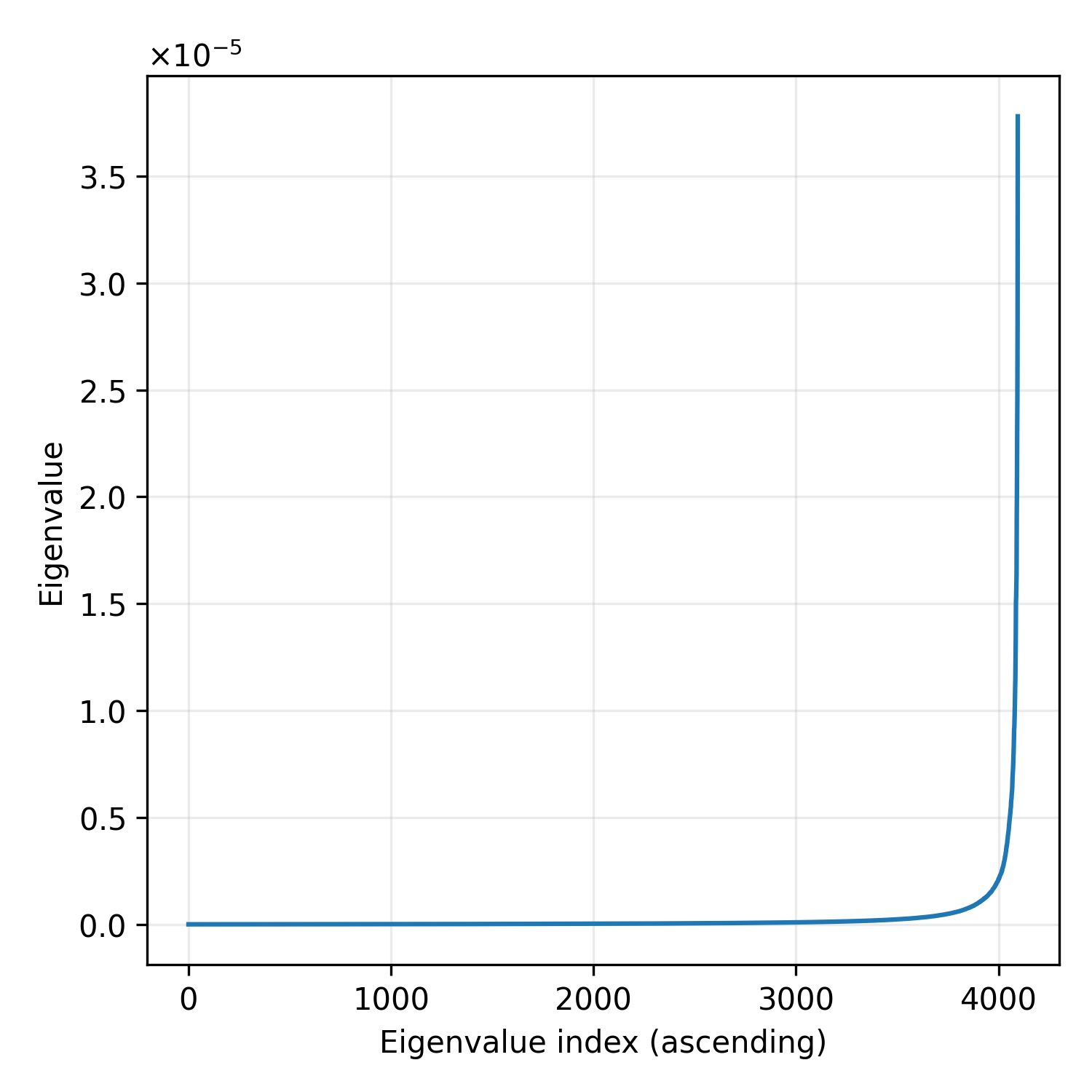} &
        \includegraphics[width=.235\columnwidth]{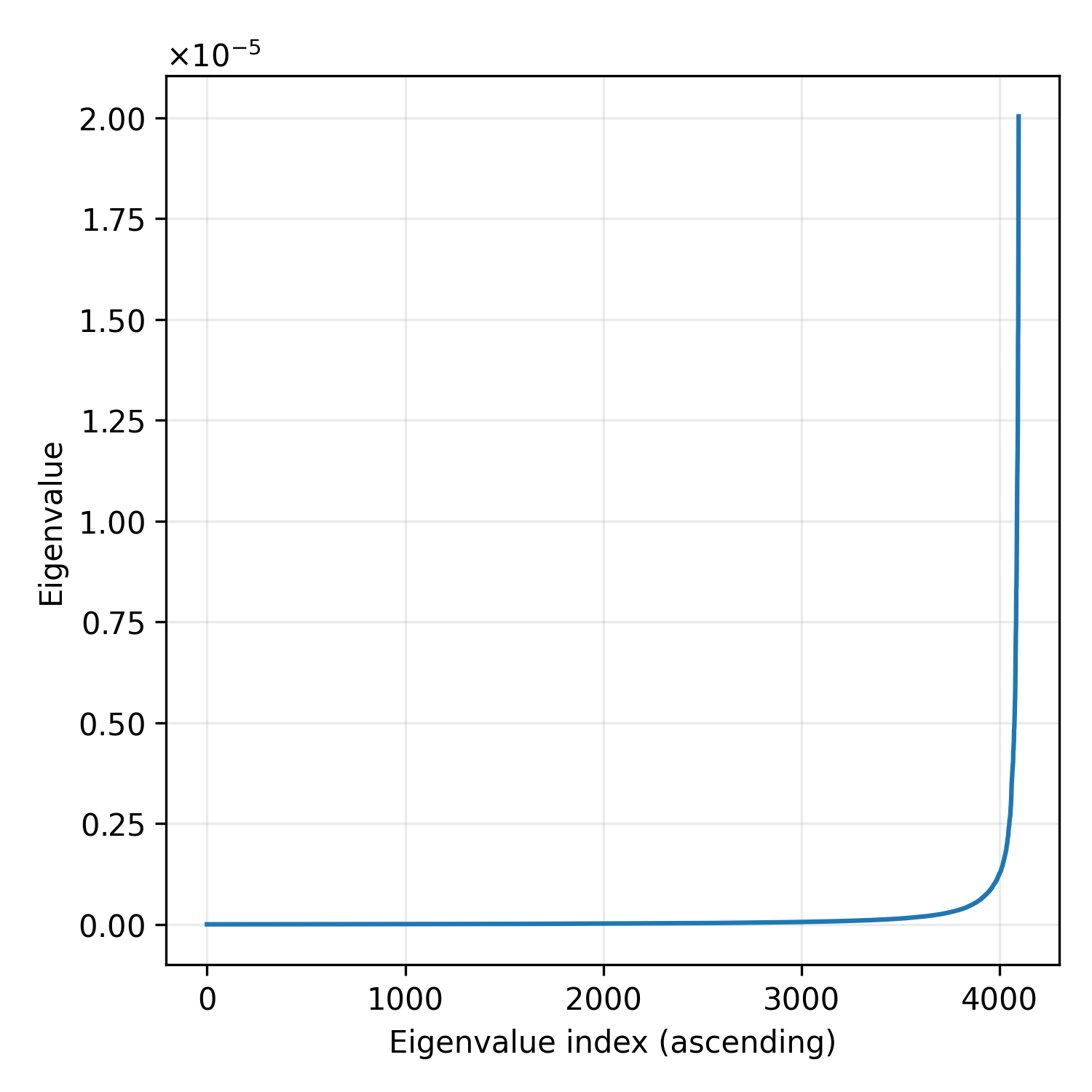} &
        \includegraphics[width=.235\columnwidth]{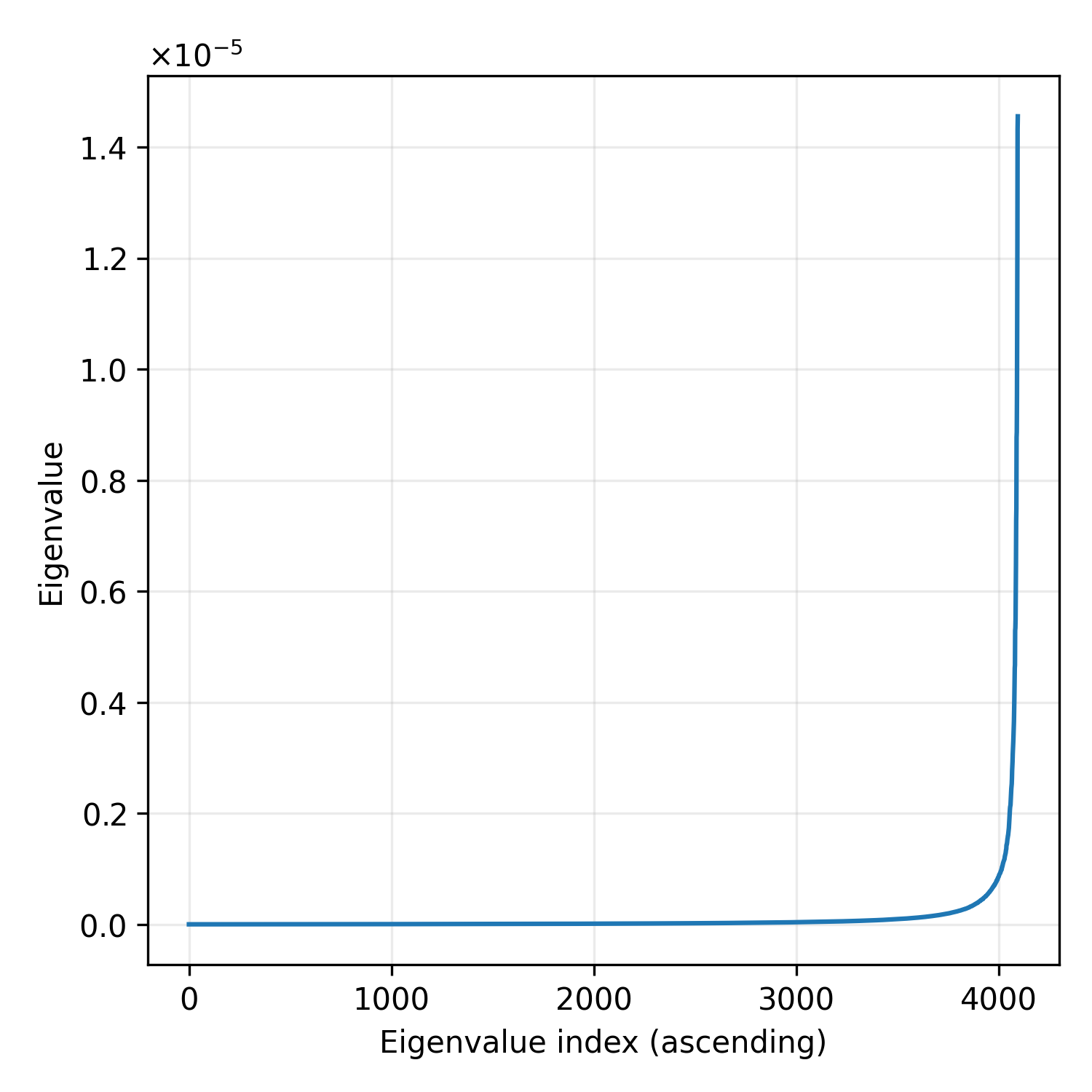} \\[-2pt]
        \includegraphics[width=.235\columnwidth]{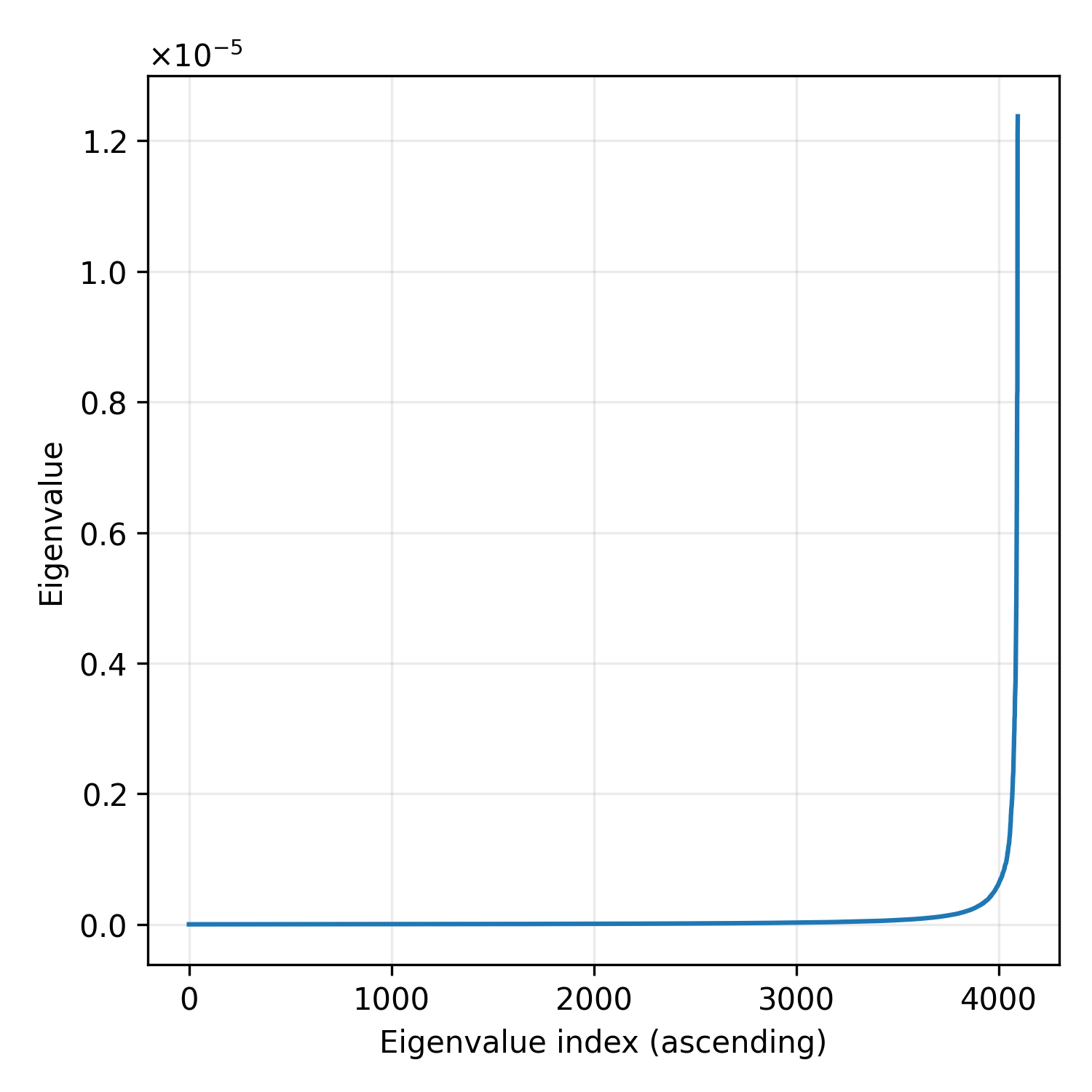} &
        \includegraphics[width=.235\columnwidth]{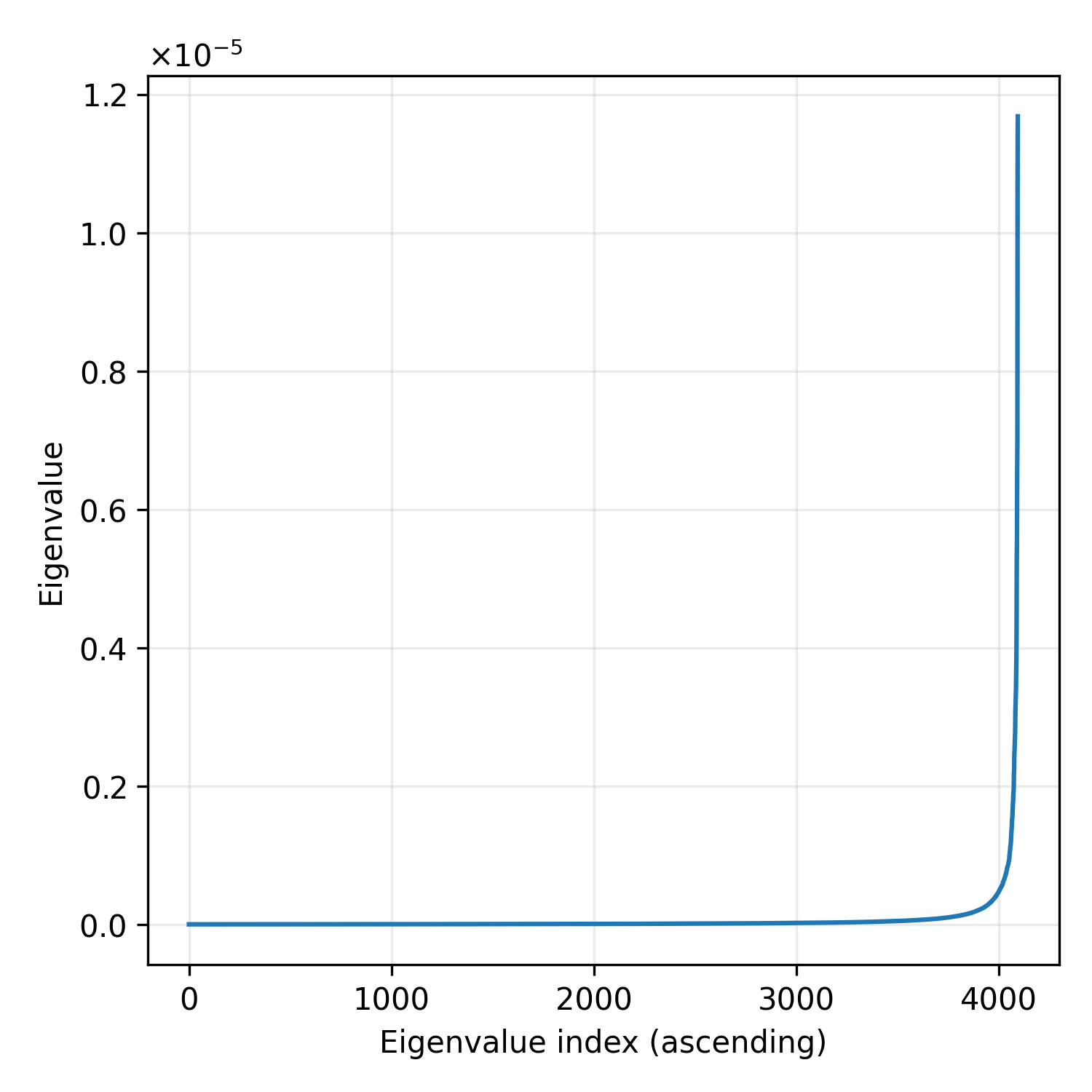} &
        \includegraphics[width=.235\columnwidth]{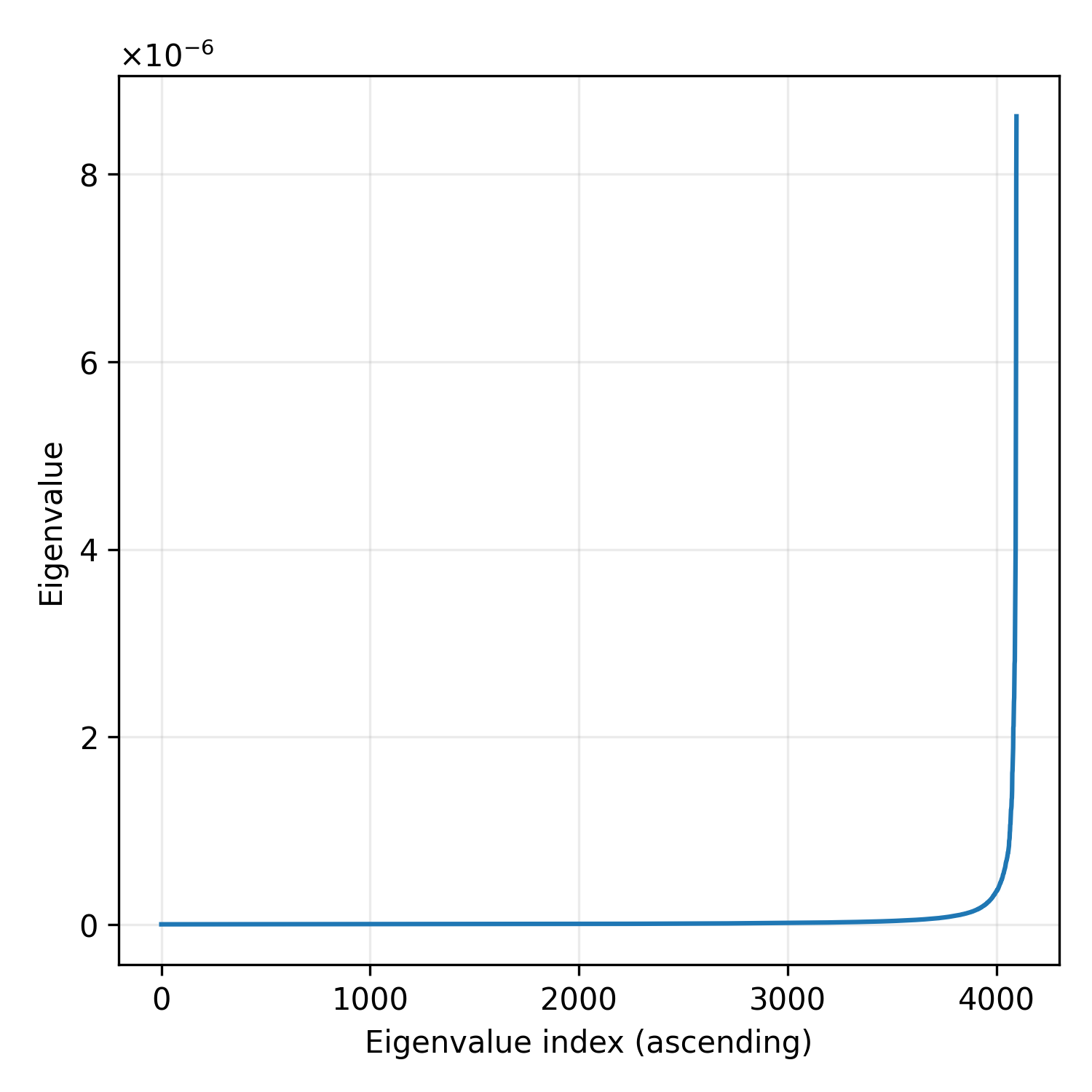} &
        \includegraphics[width=.235\columnwidth]{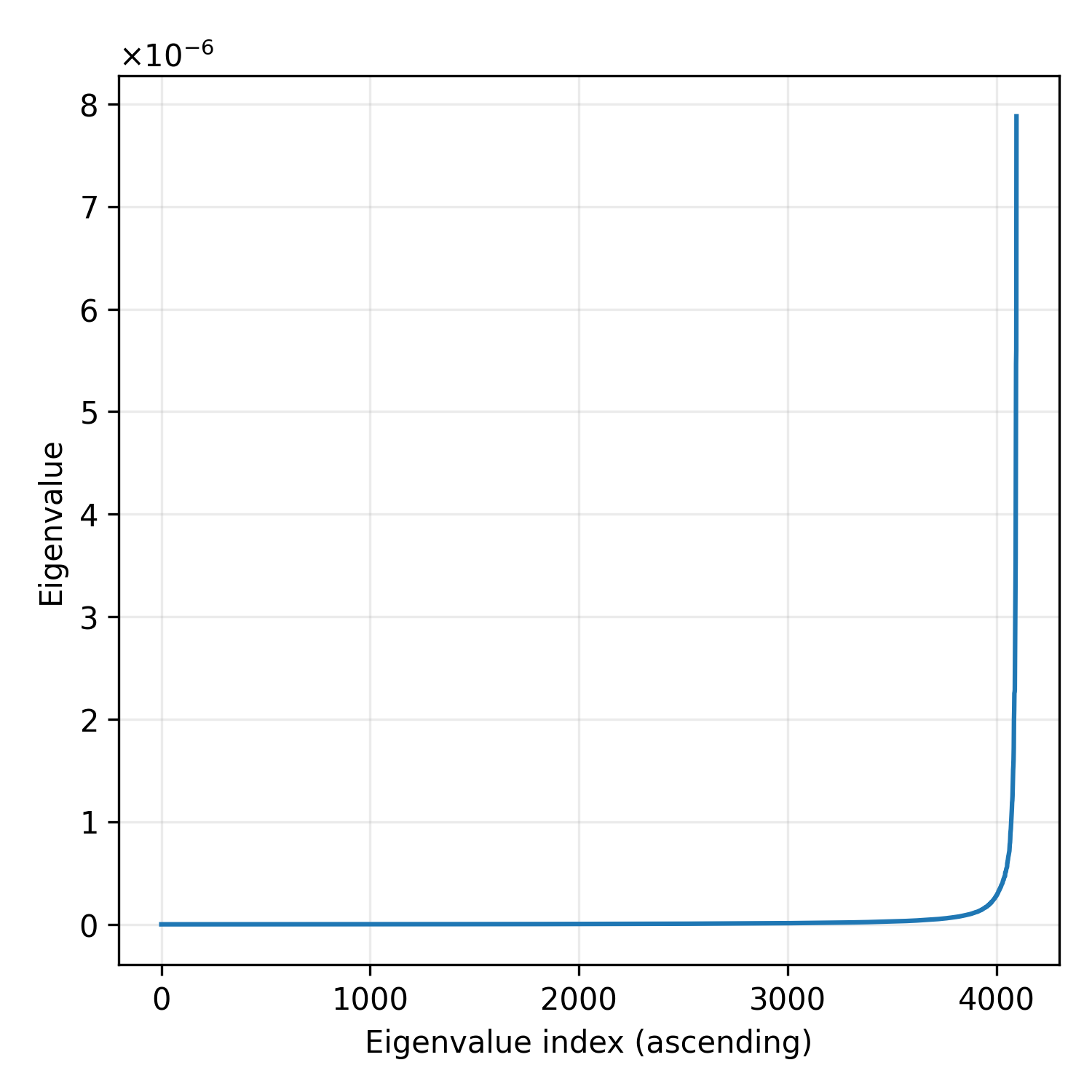} \\[-2pt]
        \includegraphics[width=.235\columnwidth]{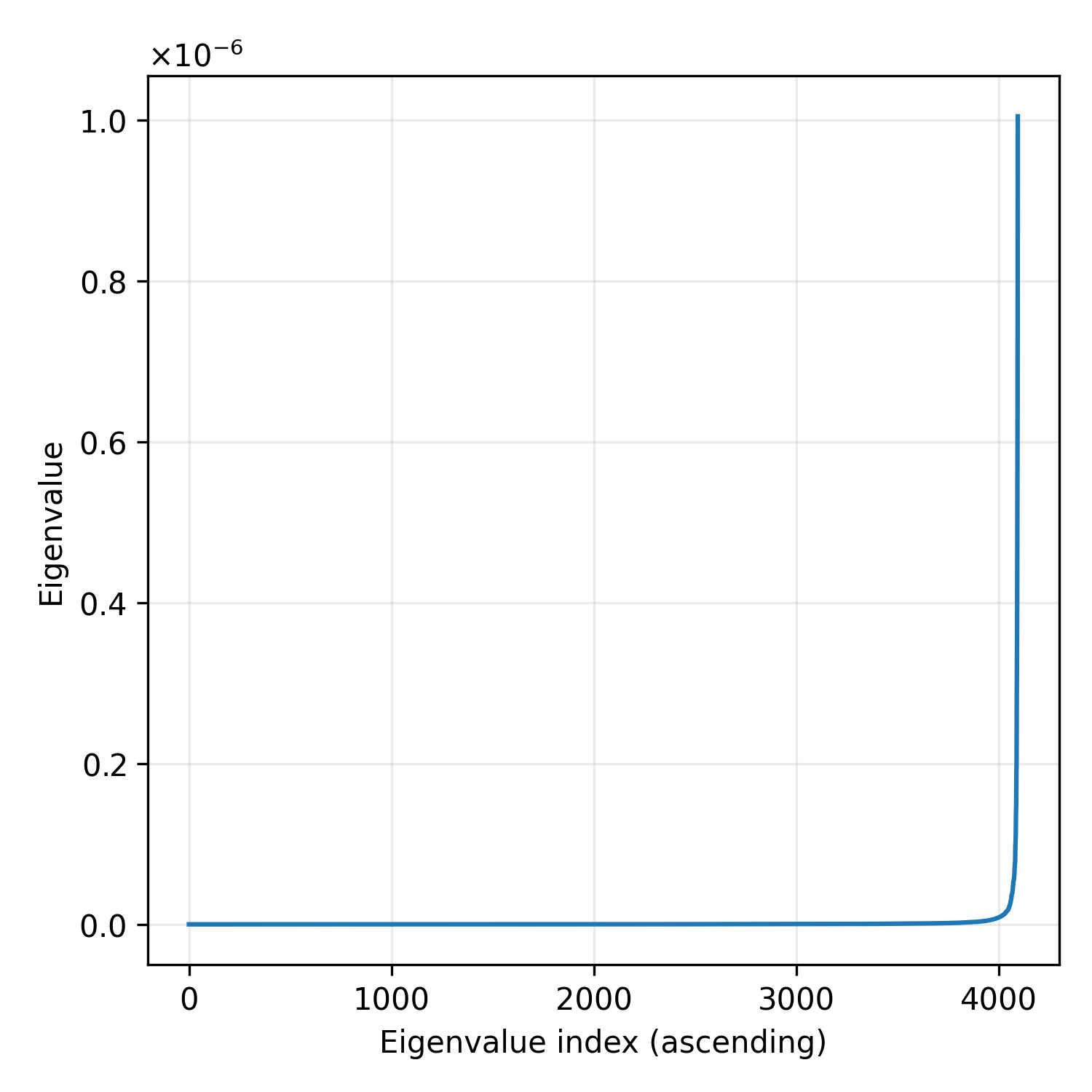} &
        \includegraphics[width=.235\columnwidth]{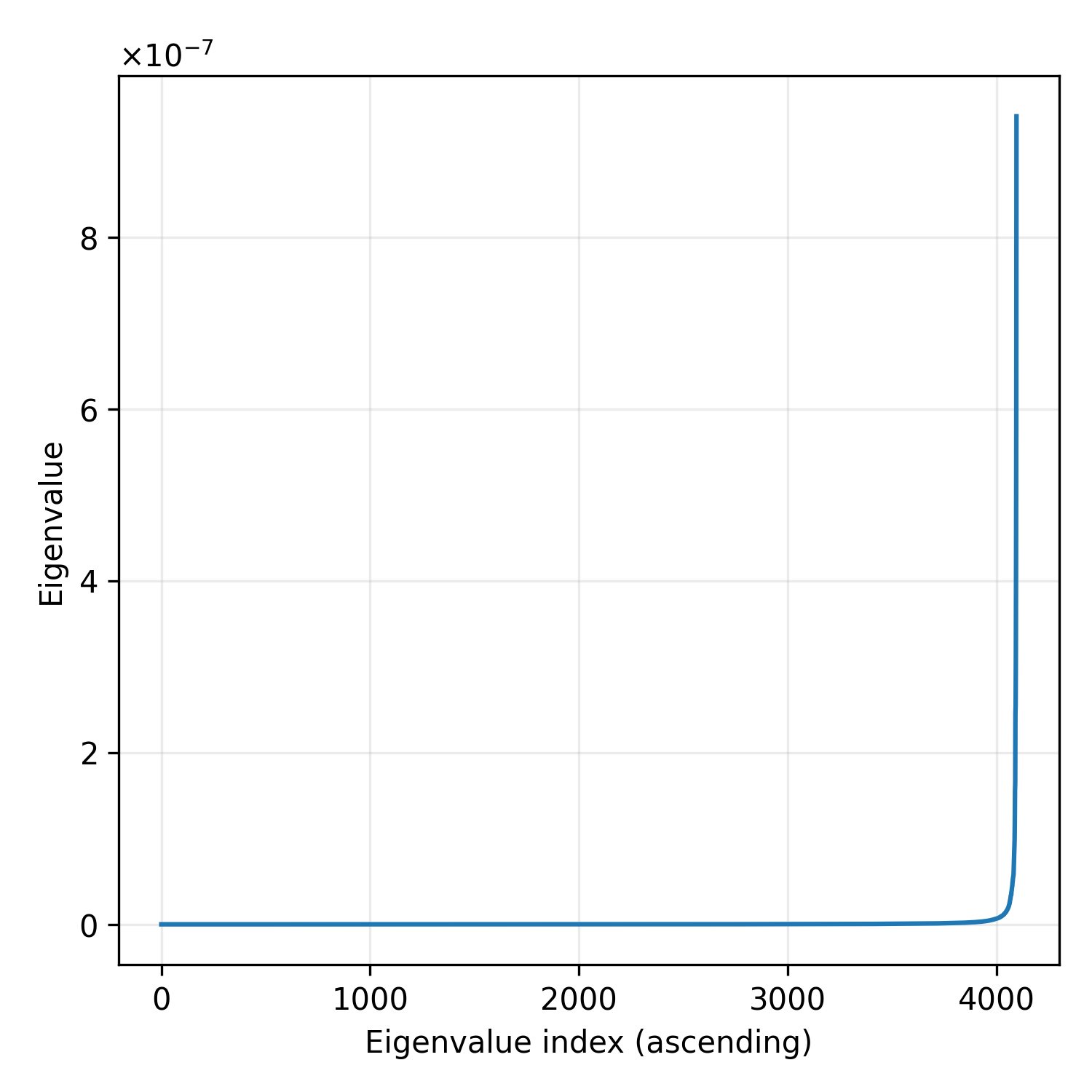} &
        \includegraphics[width=.235\columnwidth]{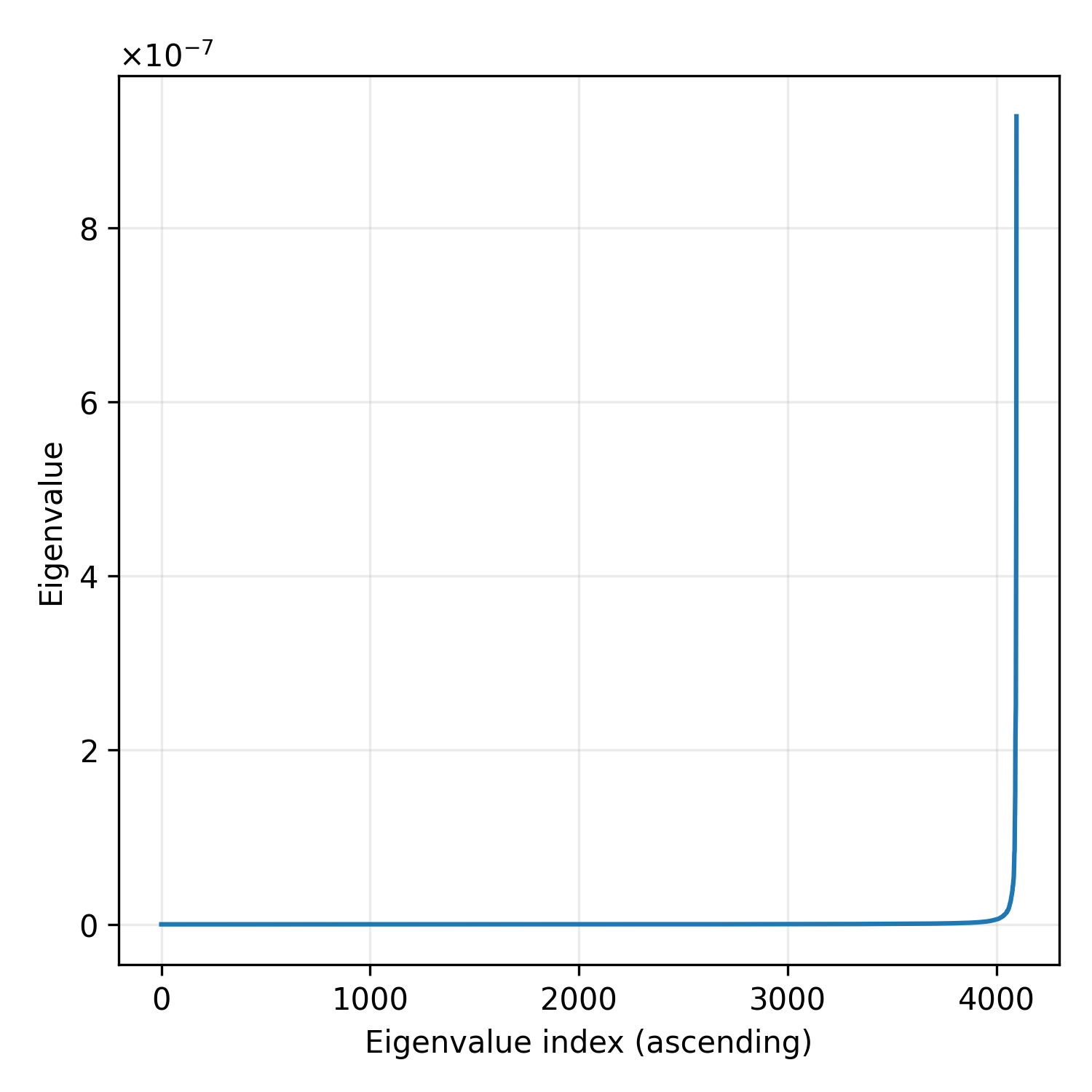} &
        \includegraphics[width=.235\columnwidth]{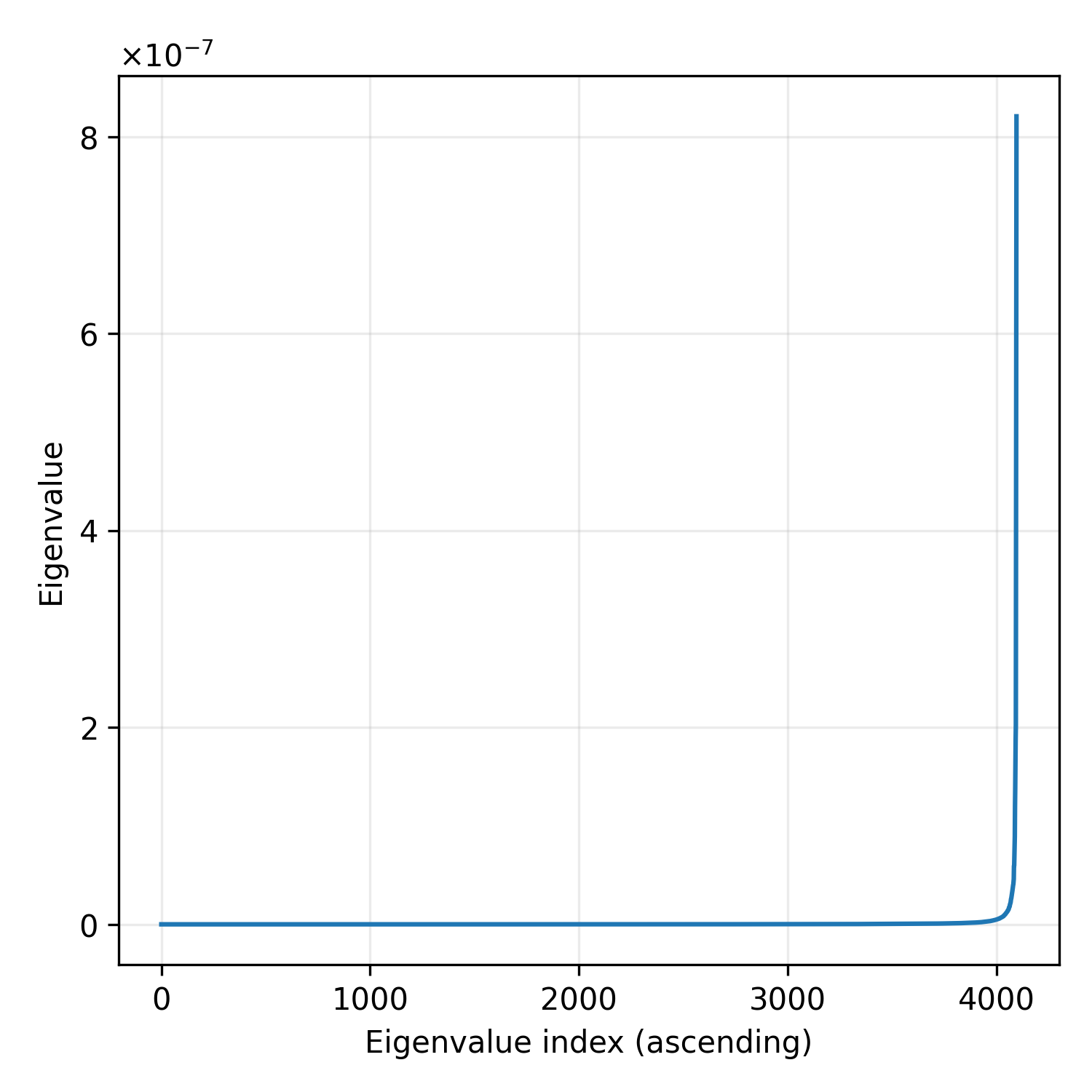} \\[-2pt]
        \includegraphics[width=.235\columnwidth]{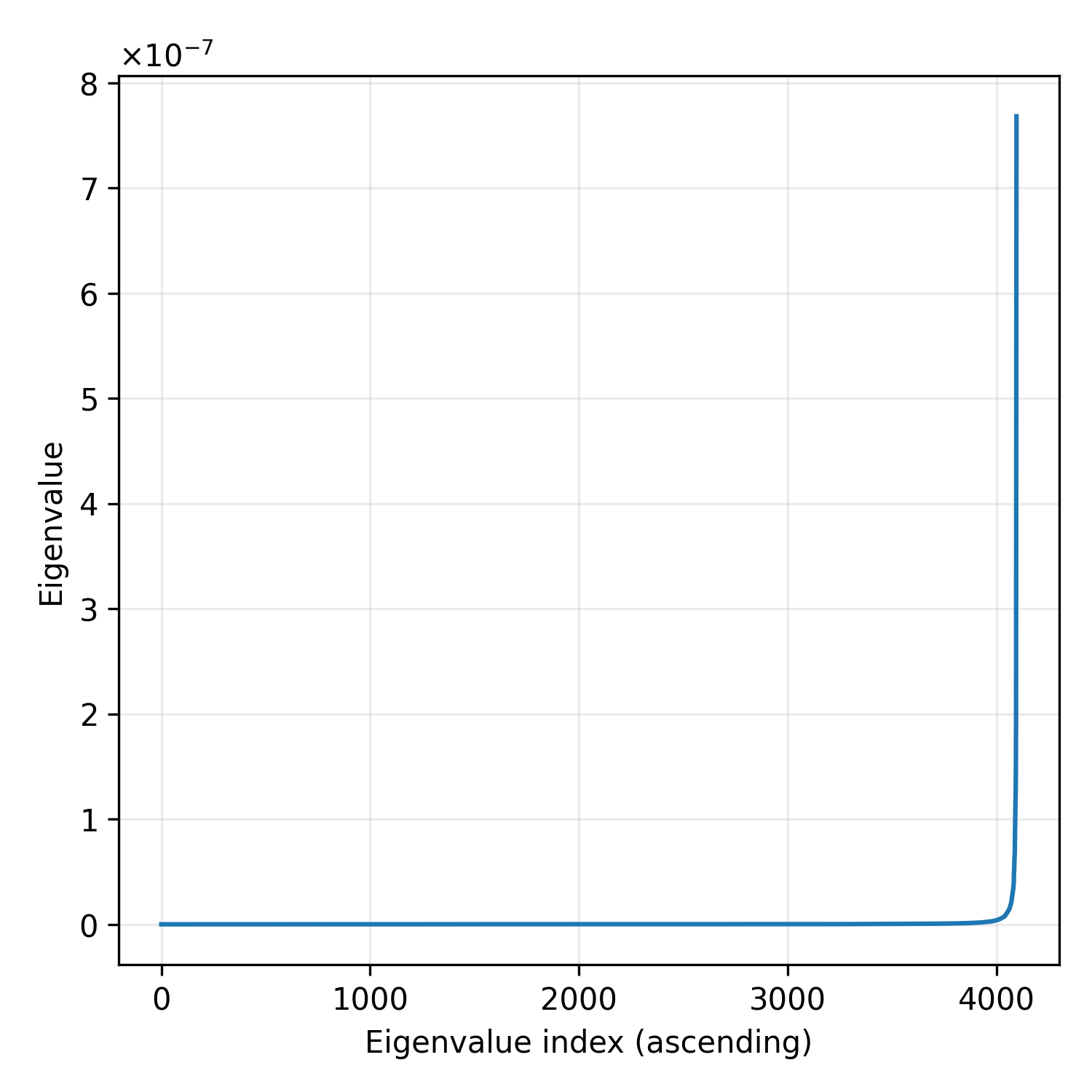} &
        \includegraphics[width=.235\columnwidth]{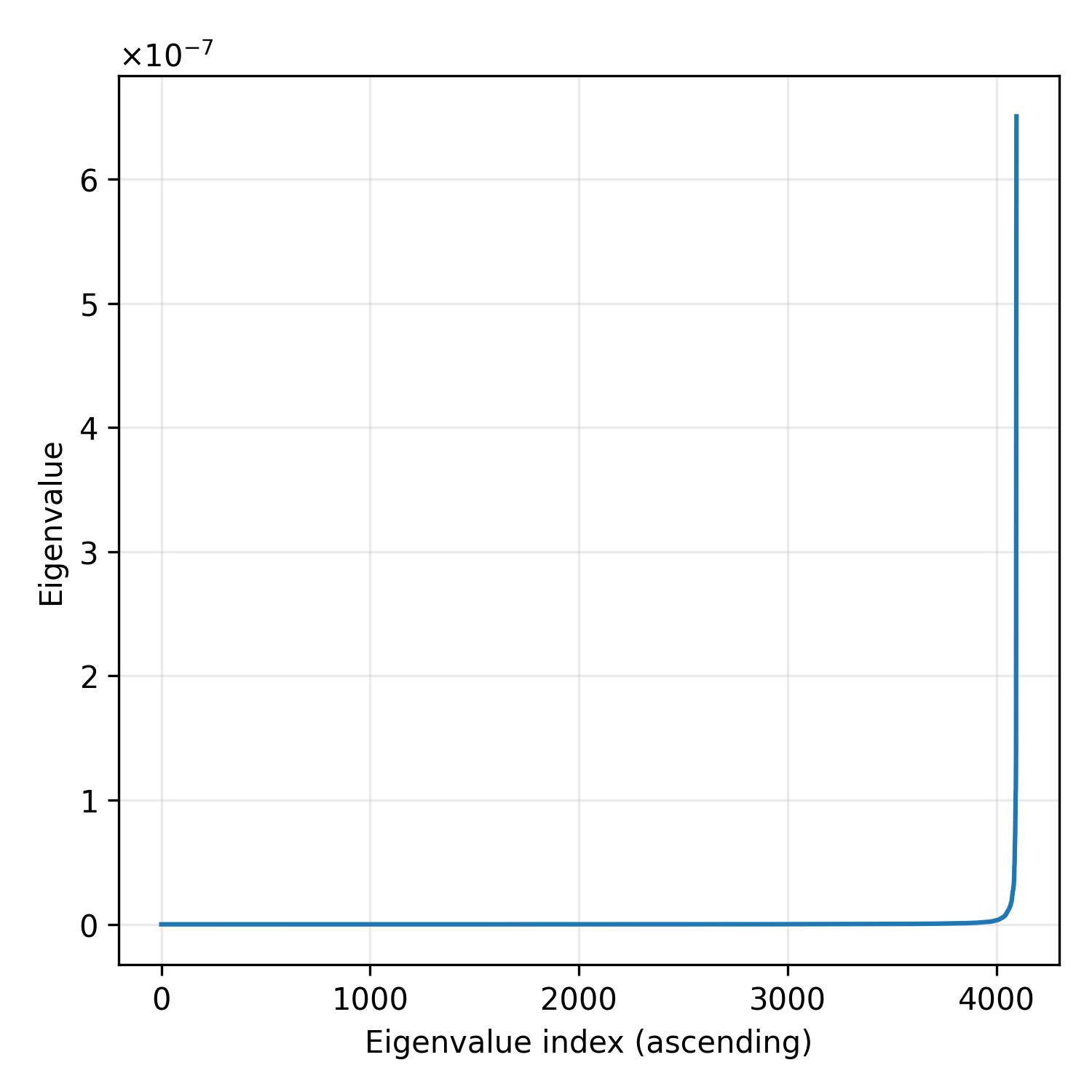} &
        \includegraphics[width=.235\columnwidth]{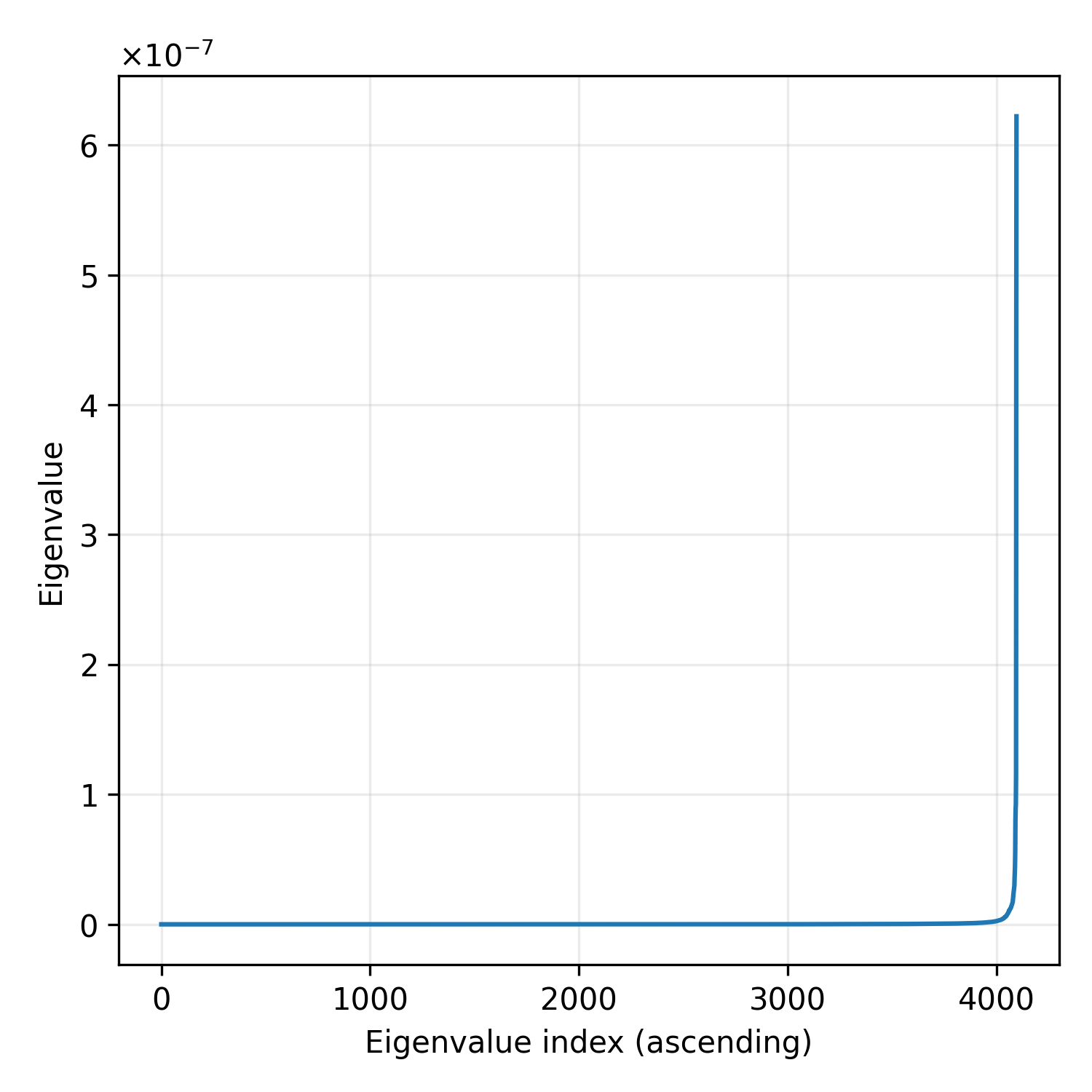} &
        \includegraphics[width=.235\columnwidth]{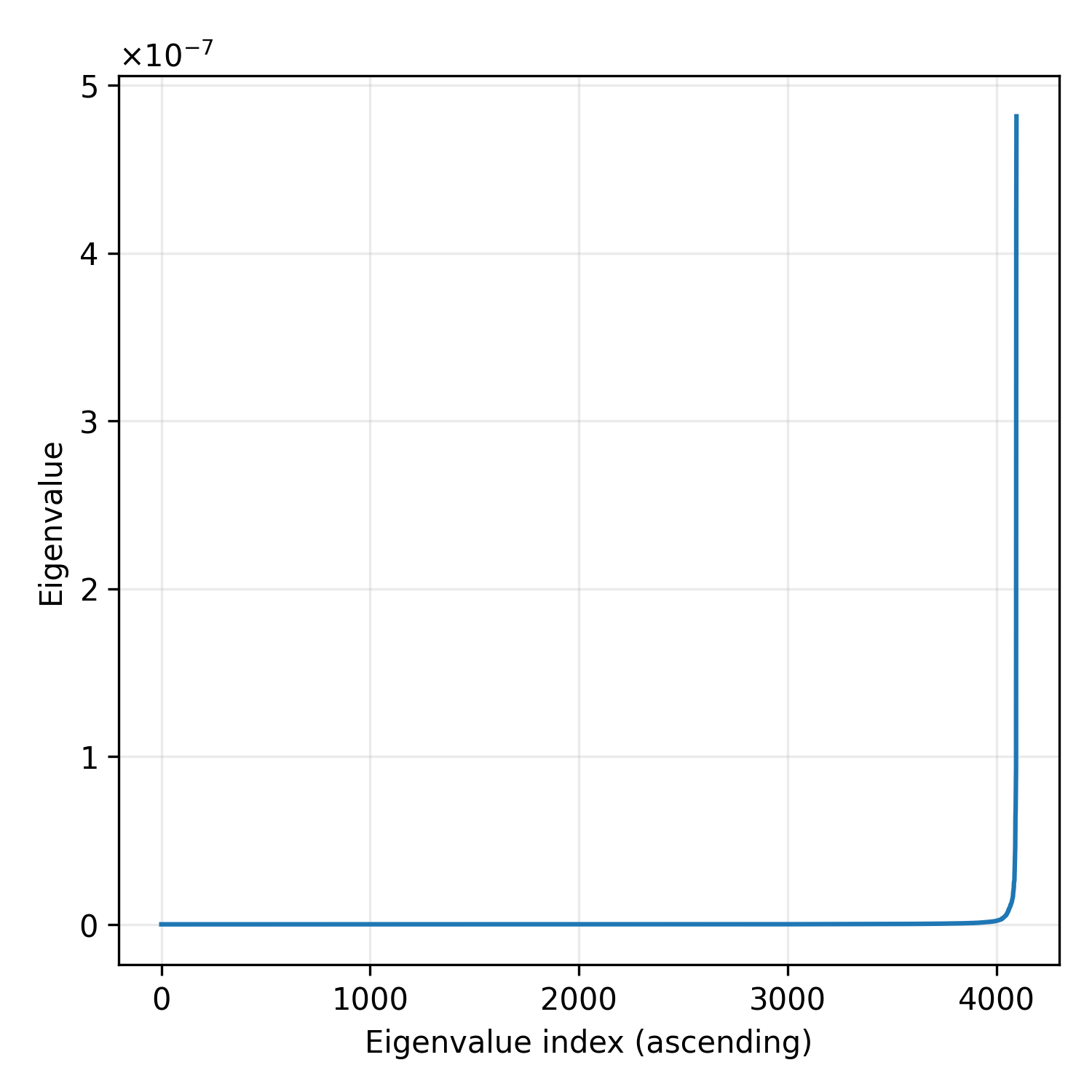} \\[-2pt]
        \includegraphics[width=.235\columnwidth]{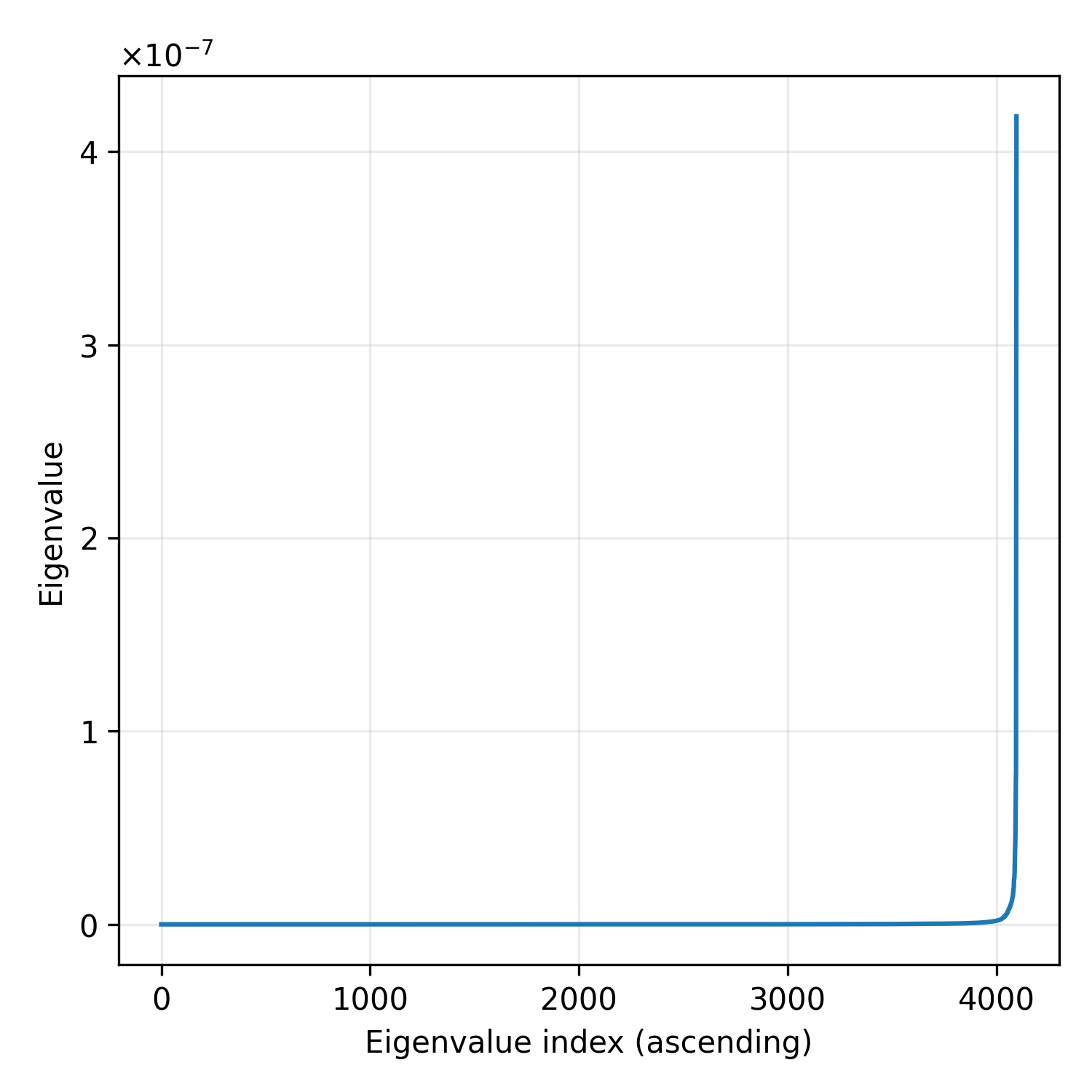} &
        \includegraphics[width=.235\columnwidth]{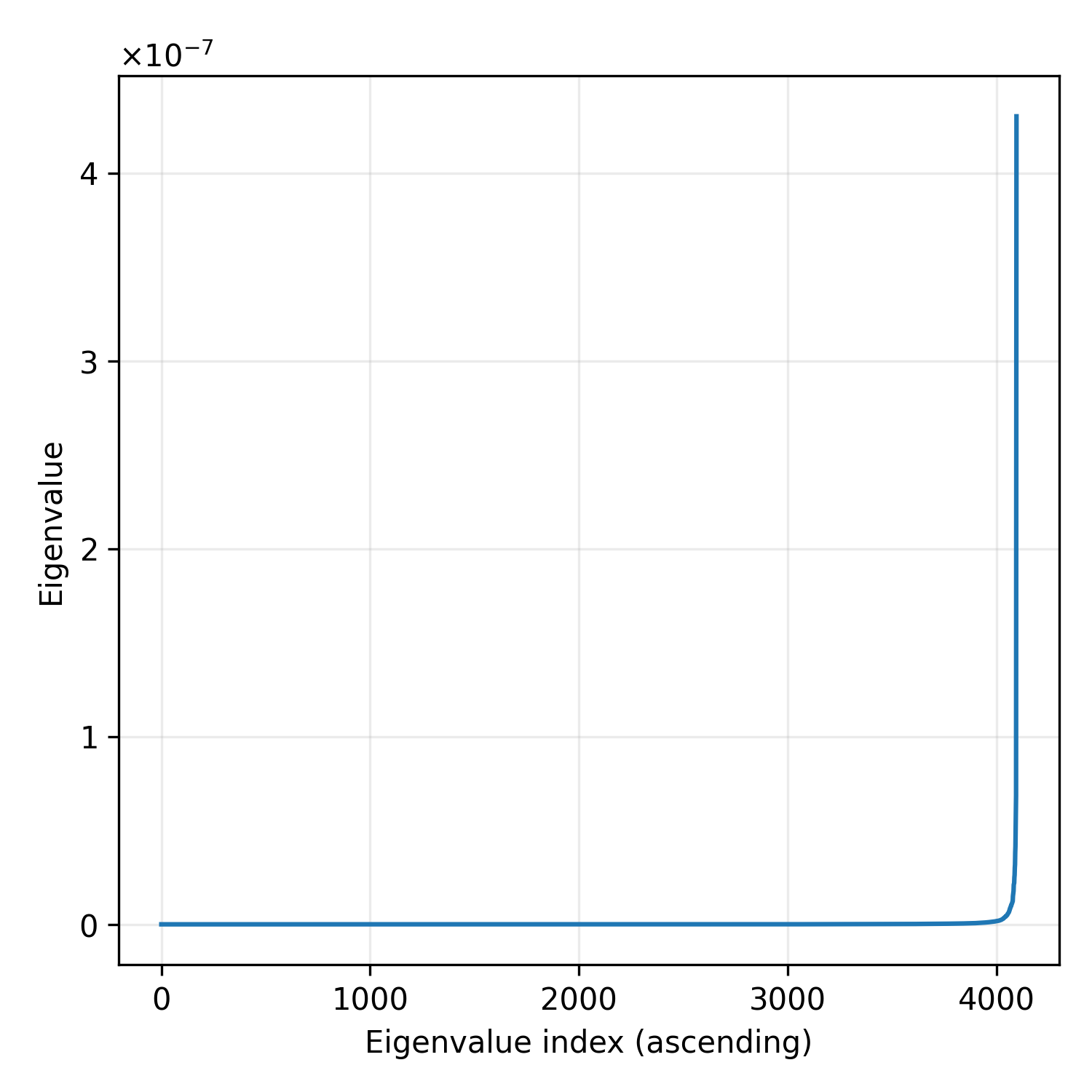} &
        \includegraphics[width=.235\columnwidth]{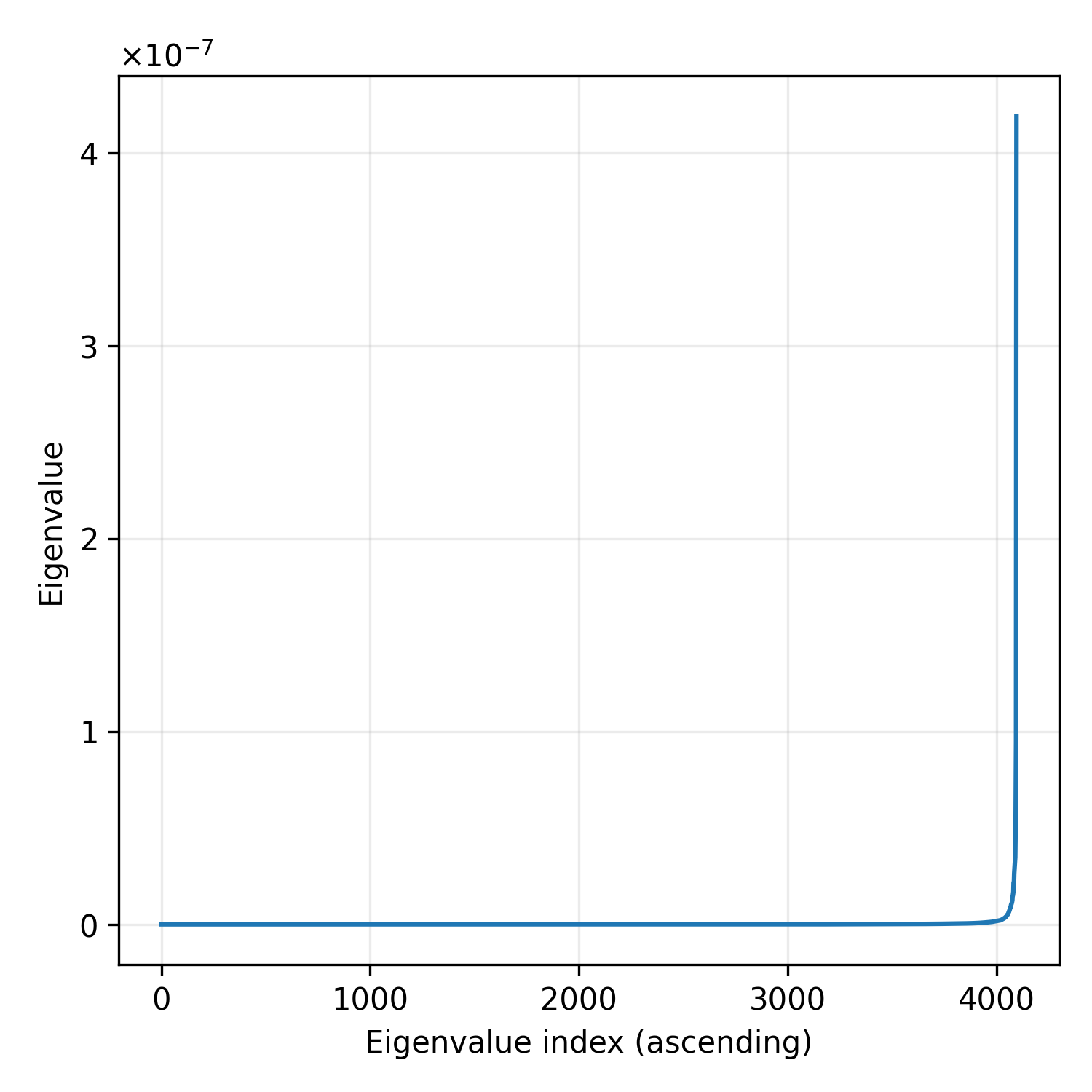} &
        \includegraphics[width=.235\columnwidth]{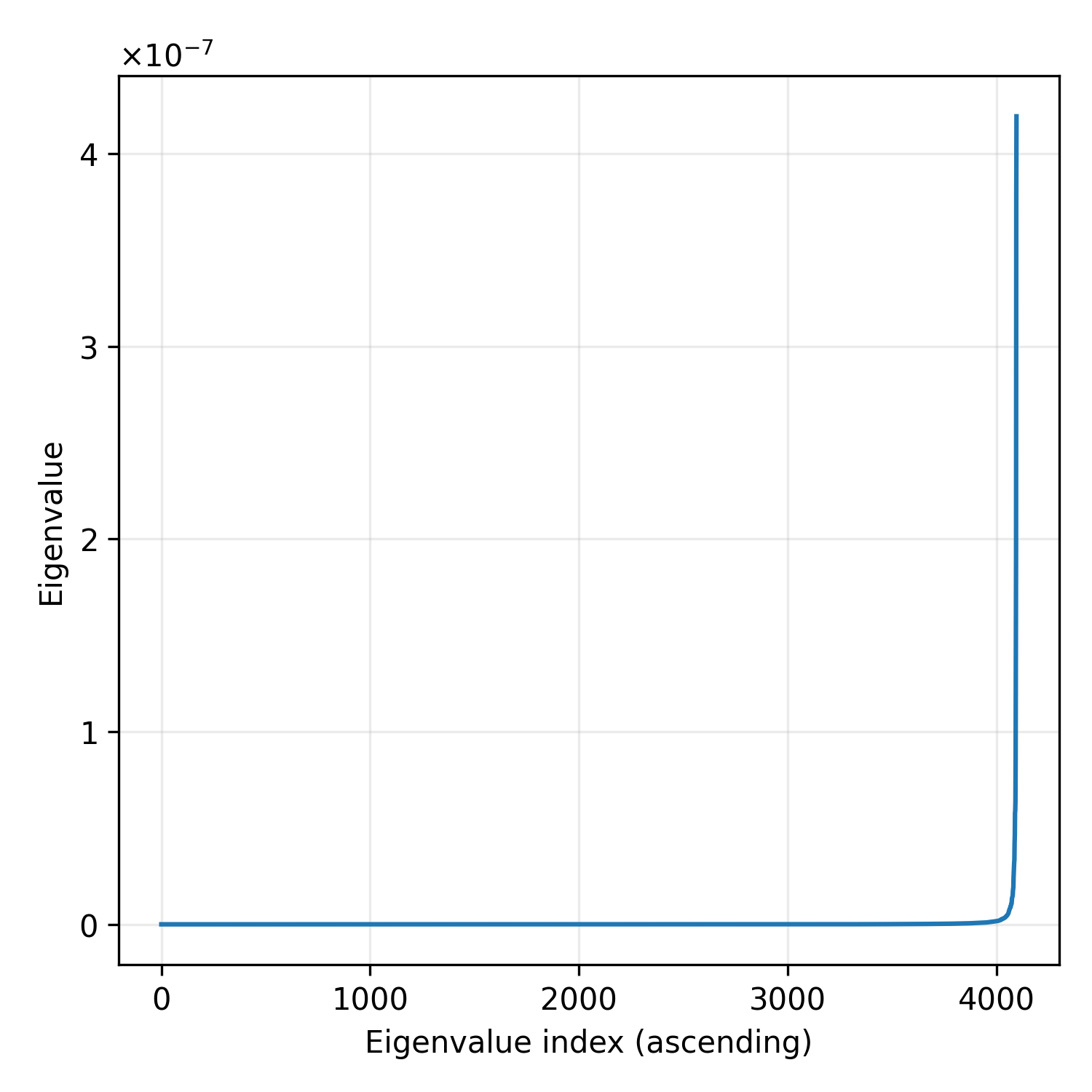} \\[-2pt]
    \end{tabular}
    \caption{
Eigenvalue spectra of the empirical Fisher matrices(excluding layers 8–19).}
    \label{fig:32-panels}
\end{figure}

\subsection{Eigendecomposition of the empirical Fisher matrix}
After eigendecomposing the empirical Fisher matrix, we sort its eigenvalues in ascending order to examine the concentration of its spectral energy. The results for each layer, excluding layers 8–19, are shown in \cref{fig:32-panels}. Across all shown layers, the eigenvalues remain near zero for most directions and rise sharply only for a small number of directions at the right end of the spectrum. This concentration suggests that a few high-eigenvalue directions account for most of the second-order sensitivity, supporting the use of the top \(r\ll C\) eigenvectors to construct the sensitive subspace. The rise becomes steeper in deeper layers, suggesting that sensitivity is concentrated in even fewer directions.

\end{document}